\pdfoutput=1

\documentclass[11pt,table,dvipsnames]{article}

\usepackage{latex/acl}

\usepackage{times}
\usepackage{latexsym}

\usepackage[T1]{fontenc}

\usepackage[utf8]{inputenc}

\usepackage{inconsolata}
\usepackage{microtype}
\usepackage{CJKutf8}
\usepackage{graphicx}
\usepackage{bigstrut}
\usepackage{amsmath}
\usepackage{multirow, makecell, caption}
\usepackage{floatrow}
\newfloatcommand{capbtabbox}{table}[][\FBwidth]
\usepackage[normalem]{ulem}
\usepackage{amssymb}
\usepackage{amsfonts}
\usepackage{multicol}
\usepackage{tipa}
\usepackage{bm}
\usepackage[ruled,linesnumbered]{algorithm2e}
\usepackage{paralist}
\usepackage{IEEEtrantools}
\usepackage[switch]{lineno}
\usepackage{enumitem}
\usepackage{arydshln}
\usepackage{verbatim}
\usepackage{pifont}
\usepackage{csquotes}
\usepackage{xspace}
\usepackage{mdwlist}
\usepackage{subfigure}
\usepackage{array}
\usepackage{colortbl}
\usepackage{dsfont}
\usepackage{url}
\usepackage{booktabs} 
\usepackage{tabularx}
\usepackage{bbm}
\usepackage{pgfplots}
\usepackage{tikz}
\usepackage{soul} 
\usepackage{color,xcolor} 
\usepackage{listings}

\newcommand{\ie}{\textit{i.e.}\xspace}
\newcommand{\eg}{\textit{e.g.}\xspace}

\definecolor{cRed}{HTML}{FF0000}
\definecolor{cyellow}{RGB}{244, 227, 178}
\definecolor{cBlue}{RGB}{226, 232, 221}

\newcommand{\method}{\textsc{TimeArena}\xspace}
\definecolor{customcolor}{rgb}{0.2, 0.4196, 0.8039}
\definecolor{customcolor2}{rgb}{0.4, 0.8039, 0.2}

\newcommand{\coloredtexttt}[1]{\textcolor{customcolor}{\textbf{\texttt{#1}}}}
\newcommand{\coloredtextttobj}[1]{\textcolor{customcolor2}{\textbf{\texttt{#1}}}}

\makeatletter
\def\adl@drawiv#1#2#3{%
        \hskip.5\tabcolsep
        \xleaders#3{#2.5\@tempdimb #1{1}#2.5\@tempdimb}%
                #2\z@ plus1fil minus1fil\relax
        \hskip.5\tabcolsep}
\newcommand{\cdashlinelr}[1]{%
  \noalign{\vskip\aboverulesep
           \global\let\@dashdrawstore\adl@draw
           \global\let\adl@draw\adl@drawiv}
  \cdashline{#1}
  \noalign{\global\let\adl@draw\@dashdrawstore
           \vskip\belowrulesep}}
\makeatother

\title{
\method: Shaping Efficient Multitasking Language Agents 
\\in a Time-Aware Simulation
}

\author{Yikai Zhang\textsuperscript{\rm $\spadesuit$},
Siyu Yuan\textsuperscript{\rm $\spadesuit$},
Caiyu Hu\textsuperscript{\rm $\spadesuit$},\\
\bf Kyle Richardson\textsuperscript{\rm $\heartsuit$},
Yanghua Xiao\textsuperscript{\rm $\spadesuit$}\thanks{Corresponding authors.},
Jiangjie Chen\textsuperscript{\rm $\spadesuit$}\footnotemark[1]
\\
\textsuperscript{\rm $\spadesuit$}Fudan University\quad
\textsuperscript{\rm $\heartsuit$}Allen Institute for AI\\
\texttt{\{ykzhang22,syyuan21,cyhu24\}@m.fudan.edu.cn},\\
\texttt{kyler@allenai.org},
\texttt{\{shawyh,jjchen19\}@fudan.edu.cn}}

\pgfplotsset{compat=1.18}

\begin{document}
\maketitle
\begin{abstract}
Despite remarkable advancements in emulating human-like behavior through Large Language Models (LLMs), current textual simulations do not adequately address the notion of time.
To this end, we introduce \method, a novel textual simulated environment that incorporates complex temporal dynamics and constraints that better reflect real-life planning scenarios.
In \method, agents are asked to complete multiple tasks as soon as possible, allowing for parallel processing to save time. 
We implement the dependency between actions, the time duration for each action, and the occupancy of the agent and the objects in the environment.
\method grounds to 30 real-world tasks in cooking, household activities, and laboratory work.
We conduct extensive experiments with various state-of-the-art LLMs using \method.
Our findings reveal that even the most powerful models, e.g., GPT-4, still lag behind humans in effective multitasking, underscoring the need for enhanced temporal awareness in the development of language agents.\footnote{Project page: \url{https://time-arena.github.io}.}
  
\end{abstract}

\section{Introduction}
\label{sec:intro}
Large language models (LLMs)~\cite{openai2022chatgpt,openai2023gpt4,geminiteam2023gemini,Hugo2023LLaMa,Hugo2023LLaMa2} have enabled the development of language agents (\textit{a.k.a.} LLM-based agents), which aim to simulate human behaviors in real-world scenarios through their reasoning and planning capabilites~\cite{wu2023smartplay,liu2023agentbench,gong2023mindagent,akata2023playing}.
However, planning in the real world involves temporal and resource constraints~\cite{russell2010artificial}, which are rarely implemented in most textual simulations for LLMs and language agents~\cite{wang-etal-2022-scienceworld,park2023generative}.

\definecolor{mygray}{RGB}{233,233,233}

\begin{figure}[t]
    \centering
    \includegraphics[width=\linewidth]{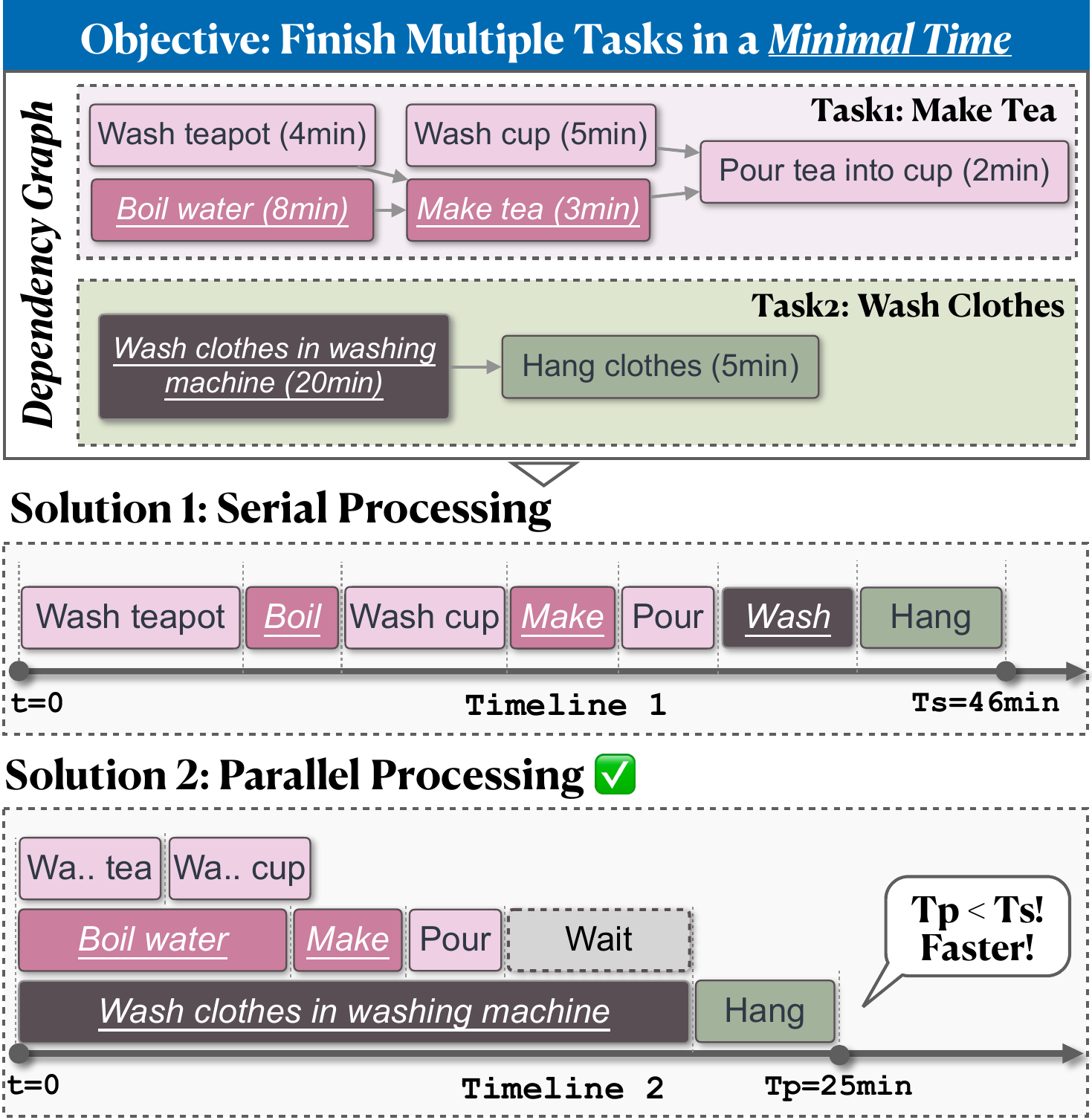}
    \caption{An example illustrating multitasking with temporal constraints in \method.
    The completion of tasks requires actions in a predetermined dependency and order. 
    \underline{Underlined actions} do not occupy the agent, allowing other actions to be processed by the agent simultaneously. 
    The \colorbox{mygray}{Wait} action skips the current time step, meaning the agent is idle.
    }
    \label{fig:front}
\end{figure}

The integration of time in simulated environments challenges agents to navigate and align with human-like efficient multitasking skills.
Such a simulation requires the agent to consider the following three factors:
\begin{inparaenum}[1)]
    \item \textbf{Time Duration and Dependency:} Actions will have durations upon dependencies, requiring agents to strategize and prioritize based on time constraints and task completion progress.
    \item \textbf{Agent Occupancy:} Agents will be occupied by certain actions thus they might be unable to perform other actions at the same time.
    \item \textbf{Object Occupancy:} 
    Some objects might be occupied for some time, and agents must use available objects in the environment for the tasks. 
\end{inparaenum}
These factors are common in real-life but are seldom addressed by current textual simulations.

To help illustrate,
Figure~\ref{fig:front} shows an example of completing the task of ``make tea'' (Task 1) and ``wash clothes'' (Task 2).
The actions of each task might depend on previous actions (\eg, agents must \coloredtexttt{boil water} before \coloredtexttt{make tea}.), and each action takes a specific duration in time (\eg, \coloredtexttt{wash cup} takes 5 minutes).
In particular, some actions let agents be idle, allowing agents to carry out other actions.
For example, \coloredtexttt{wash clothes in washing machine} allows agents to perform other actions at the same time.
Moreover, actions temporarily occupy objects, making them unavailable for other actions and hindering parallel processing. 
For example, \coloredtexttt{boil water} occupies the \coloredtexttt{pot}, delaying other actions like \coloredtexttt{cook soup} until it is available.
When no action is currently available for the agent, the only option is to wait. 
For example, in Solution 2, the agent must wait for the completion of \coloredtexttt{wash clothes in washing machine}, before \coloredtexttt{hang clothes}.

In this work, we introduce \method, a textual simulated environment featuring 30 real-world involving cooking, household activities, and laboratory work.
\method is the first textual simulation to evaluate language agents on multitasking efficiency.
Specifically, we incorporate the time duration of each action and set two types of actions based on agent occupancy.
One type occupies agents (\eg, \coloredtexttt{wash cup}) and another lets agents be idle (\eg, \coloredtexttt{boil water}). 
Additionally, we simulate resource competition by implementing object occupancy, i.e., an object used for one task cannot be simultaneously used for another, which is common in parallel processing. 
Therefore, agents must focus on parallel processing, taking into account the occupancy of agents and objects, to minimize time consumption.
We design four metrics in \method to evaluate the average progress, completion speed, task completion rate and average completion time.
These metrics help to assess and analyze the efficient multitasking capabilities of language agents. 
Our comprehensive evaluation of 7 LLMs on \method shows that current language agents struggle in efficient multitasking.
Even the most powerful LLM, GPT-4, still faces challenges in parallel processing.

In summary, our contributions are as follows:
\begin{itemize}
    \item To the best of our knowledge, we are the first to explore the notion of time of language agents in a textual environment, which is important for more realistic simulation.
    \item We create \method, a novel text-based simulated environment consisting of 30 tasks, where LLMs can complete multiple tasks in parallel.
    \item Using \method, we conduct rich experiments to evaluate the efficient multitasking capabilities of language agents. Our results demonstrate that efficient multitasking in \method poses a significant challenge for current LLMs and language agents.
\end{itemize}

\section{Related Work}
\label{sec:related}
\paragraph{Simulation-based Evaluation For language Agents}
With the great success of LLMs~\cite{ouyang2022training,openai2022chatgpt,openai2023gpt4,geminiteam2023gemini,Hugo2023LLaMa,Hugo2023LLaMa2}, recent works have shifted the focus from traditional NLP tasks to explore language agents in simulation environments that mimic real-world scenarios~\cite{wu2023smartplay,liu2023agentbench,gong2023mindagent,akata2023playing}.
These simulation environments can be divided into two categories:
1) Social Simulations~\cite{park2023generative,mukobi2023welfare,zhou2023sotopia}, which aim to evaluate the behaviors of language agents in some social scenarios;
2) Problem-solving simulations, which are created based on competitive games~\cite{wang2023avalons,xu2023language,chen2023put} or cooperation games~\cite{chen2023agentverse,zhang2023building,agashe2023evaluating} and scientific scenarios~\cite{wang-etal-2022-scienceworld}. 
These simulations mainly test the planning and reasoning abilities of language agents and need agents to solve specific problems within dynamic and evolving environments.
In this paper, we focus on problem-solving simulations to investigate the efficient multitasking capabilities of language agents.

\begin{figure*}[t]
    \centering
    \includegraphics[width=\linewidth]{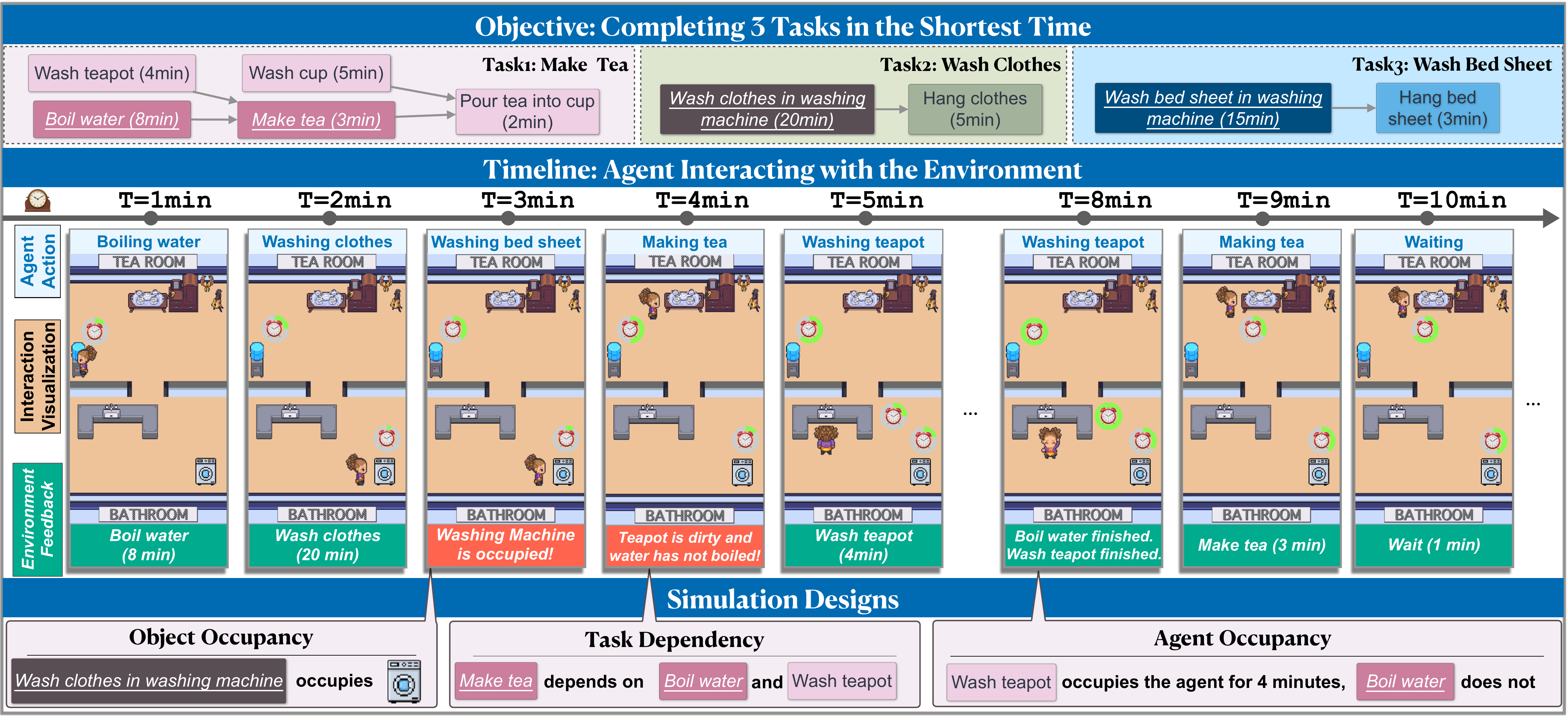}
    \caption{An overview of \method, with a multitasking example that shows our designs of the simulation. 
    \method first sets an objective for the agent, and then the agent interacts with \method over time, with the design of task dependency, object occupancy, and agent occupancy.}
    \label{fig:TimeArena}
\end{figure*}

\paragraph{Language Planning}
Language planning aims to decompose a complex task into steps~\cite{schank1975scripts,schank2013scripts}.
Early studies mainly endow the planning capabilities of LMs through training them on specific planning datasets~\cite{peng-etal-2018-deep,hua-etal-2019-argument-generation,kong-etal-2021-stylized}, which exhibits poor generalization.
Recent studies have identified that LLMs can effectively decompose tasks into procedural steps~\cite{olmo2021gpt3,lu2022neuro,ruan2023tptu,TaPA,song2023llmplanner,wang2023describe,yuan-etal-2023-distilling,shen2023taskbench}.
However, existing work mostly focuses on planning the logical structure of actions, where these actions lack specified time durations and do not accommodate the possibility of agents performing multiple actions concurrently.
Multitask planning with parallel processing in dynamic environments still remains under-studied.

\paragraph{Temporal Reasoning}
Numerous research efforts address diverse challenges in temporal reasoning.
Temporal reasoning involves comprehending, structuring, and interpreting events, actions, and states through the lens of time~\cite{allen1991planning,vila1994survey,stock1998spatial}.
Previous studies in temporal reasoning focus on temporal relation extraction~\cite{miller-etal-2015-extracting,vashishtha-etal-2019-fine,mathur-etal-2021-timers,wang-etal-2023-extracting}, event
temporal reasoning~\cite{zhou2020temporal,qin-etal-2021-timedial,10.1162/tacl_a_00459,mathur-etal-2022-doctime,yang-etal-2023-upon,wang2023tram} and explore the temporal reasoning capability of LLMs with several contemporary time-sensitive QA datasets~\cite{zhang-choi-2021-situatedqa,shang-etal-2022-improving,tan-etal-2023-towards}.
Distinguished from other datasets and benchmarks~\cite{chu2023timebench}, our \method creates a dynamic and interactive simulated environment.
Agents need to interact with \method over time and decide the relationship of actions in the evolving environment.

\section{\method}
\label{sec:method}
We create \method, a textual simulated environment to evaluate the efficient multitasking capabilities of language agents.
To help illustrate, we first show an overview and an example run of how an agent interacts with the \method environment ($\mathsection$~\ref{sec3:overview}), and then describe the design of the simulation environment in more detail ($\mathsection$~\ref{sec3:components}-\ref{sec3:feedback}).

\subsection{Overview of \method}
\label{sec3:overview}

\method challenges agents to complete multiple tasks strategically in the shortest possible time. 
This simulation emphasizes the importance of understanding, performing, and optimizing actions within a constrained timeframe, mirroring practical scenarios involving time management.

Central to \method are \textbf{Tasks}, \textbf{Objects}, and \textbf{Actions}. \textbf{Tasks} define the objectives for the agents, \textbf{Objects} represent elements in the environment that agents will encounter and interact with, and \textbf{Actions} are the means to accomplish these tasks. 
Real-time feedback and scoring mechanisms are integral to the environment, assessing agent performance and adding to the simulation's complexity and realism. 
Unique features like the duration and occupancy of actions and strategic resource utilization distinguish \method from other environments.

\paragraph{An Example Run}
As in Figure~\ref{fig:TimeArena}, consider an agent tasked with ``make tea'' (Task 1), ``wash clothes'' (Task 2) and ``wash bed sheet'' (Task 3).
The agent starts by decomposing the task into actions like \coloredtexttt{boil water}.
In \method, all actions have a duration (\eg, \coloredtexttt{Boil water} needs 8 minutes.) and dependencies (\eg, At T=4min, \coloredtexttt{make tea} violates the dependency because \coloredtexttt{wash teapot} and \coloredtexttt{boil water} are not completed yet.).
The agent then interacts with objects (\eg, \coloredtexttt{wash clothes in washing machine}), which become \textbf{occupied} during the process. 
The agent can engage in non-occupied
actions simultaneously (\eg, \coloredtexttt{wash teapot}) while others (\eg, \coloredtexttt{boil water}) are in progress.
Environmental feedback guides the agent, indicating the legitimacy of actions and the completion of tasks. 
For example, if the \coloredtexttt{washing machine} is occupied, the agent adjusts its strategy. 
The agent's goal is to complete all tasks efficiently, with performance evaluated based on progress and completion time.

This dynamic interaction in \method fosters an environment where strategic planning, resource management, and adaptability are key to an agent's success.

\subsection{Components of \method}
\label{sec3:components}

\paragraph{Tasks}
In \method, We design tasks within three distinct \textit{scenarios} or simulated settings, namely, \textit{household activities}, \textit{cooking}, and \textit{laboratory work}.
Each scenario represents a specific context or environment where multitasking is an integral part of the activities involved.\footnote{Details of tasks are in Appendix~\ref{app:task}.}
For example, one can do \coloredtexttt{sweep floor} while doing \coloredtexttt{boil water}, and do \coloredtexttt{wash dishes} while doing \coloredtexttt{cook soup}.
Each scenario contains 10 tasks,
and some actions and objects are shared across multiple tasks of a scenario.
Each task requires multiple actions to be executed, which manipulates the objects in the environment for task completion.
In the beginning, \method gives a list of tasks to the agent, with a comprehensive task instruction consisting of a \textit{task description}, an \textit{action space}, and an \textit{object set}:
\begin{itemize}[noitemsep]
    \item \textbf{Task Description}: Introduces task objectives (\eg, ``Task 1: Make a dish of beef fried rice, which consists of cooked rice and fried beef.'');
    \item \textbf{Action Space}: Lists the valid actions for the tasks (\eg, \coloredtexttt{chop}, \coloredtexttt{wash});
    \item \textbf{Object Set}: Lists the available objects in the environment for the tasks (\eg, \coloredtexttt{pot}, \coloredtexttt{beaker}).
\end{itemize}
At every timestep $t$, the agent needs to generate valid actions on the objects and receive feedback from the environment.

\paragraph{Objects}
Objects are integral to completing tasks and situating within the environment.
In \method, there are 71 different objects for all the tasks.
Every task involves a list of objects, which might overlap with other tasks of the same scenario.
To mimic the resource limitation in real-world parallel processing, we introduce:
\begin{itemize}
    \item \textbf{Object Occupancy}: the state of the object involved in an action is set to be \textit{occupied}, \eg, \coloredtexttt{wash cup} will cause the object \coloredtexttt{cup} to be occupied.
    This object cannot be processed until the involved action is completed (after some time).
    Then, this object is reset as non-occupied and waits for another action.
\end{itemize}

\paragraph{Actions}

We design a total of 45 actions for all 30 tasks.
Each action consists of a detailed description (\eg, \coloredtexttt{chop OBJ}, \coloredtexttt{chop the whole item into sliced pieces.}), showing a change of states the action will cause to an object.\footnote{All the actions are listed in Apendix~\ref{app:action}.}
Different from existing text-based simulations~\cite{wang-etal-2022-scienceworld,gong2023mindagent,shridhar2020alfworld}, in our case, an action has a duration of time and may occupy the agent from performing other actions, to the passage of time.
In detail:
\begin{itemize}[noitemsep]
    \item \textbf{Action Dependency}: An action within the same task might depend on completing other actions within the same task. 
    As depicted in Figure~\ref{fig:TimeArena}, \coloredtexttt{make tea} is dependent on \coloredtexttt{wash teapot}.
    \item \textbf{Duration of Time}:
    Each action holds a timeframe in the timeline, ranging from 1 to 10 minutes.
    In practice, agents only have an educated guess of the time duration of each action until actually interacting with \method.  
    \item \textbf{Agent Occupancy}:  
    One key to parallel processing is agent occupancy, which prevents agents from performing other tasks.
    Therefore, we consider two types of actions based on agent occupancy: Type 1 action occupies the agent til completion (\eg, \coloredtexttt{wash teapot}); and Type 2 action lets agents be idle, allowing to perform other actions (\eg, \coloredtexttt{boil water}).
\end{itemize}

\subsection{The Interaction between Agent and Environment}
\label{sec3:feedback}
\paragraph{Environmental Feedback}
The feedback from a textual environment is important to simulate and implement the constraints in \method using only textual messages.
We define feedback as the response from the environment following an action by an agent.
A feedback message could be of multiple types, including:
\begin{itemize}[noitemsep]
    \item \textbf{Invalid Action}:
    An action attempt that does not match the required format, \eg, \textit{``\coloredtexttt{clean teapot} is invalid''}.
    \item \textbf{Action on Non-existing Object}:
    An action attempt that visits objects that are not in the object set, \eg, \textit{``\coloredtexttt{pan} is non-existent''}.
    \item \textbf{Wrong Action Input}:
    An action attempt that the prerequisite action has not completed (\eg, \textit{``Cannot perform action \coloredtexttt{add to} on object \coloredtexttt{shrimp}. Because \coloredtexttt{shrimp} is \coloredtexttt{raw}''.}) or has been completed (\eg, \textit{``\coloredtexttt{wash beaker} has been completed''}).
    \item \textbf{Action on Mismatched Object}:
    An action attempt that does not match the object, \eg, \textit{``You cannot perfrom \coloredtexttt{read} on \coloredtexttt{potato}.''}
    \item \textbf{Action on Occupied Object}:
    An action attempt on occupied objects, \eg, \textit{``Object \coloredtexttt{pot} is being occupied by another action''.}
\end{itemize}
Correspondingly, valid actions will trigger environmental feedback of the following types:
\begin{itemize}[noitemsep]
    \item \textbf{Action Start}:
    Avoiding previous errors,
    valid actions will receive a feedback message containing the specific performing time, marking the start of the action, \eg, \textit{You are doing \coloredtexttt{wash cup}, it will take 9 minutes}.
    \item \textbf{Action Completion}:
    When an action is completed, the environment will send a message, \eg, \coloredtexttt{cup is clean}, and reset the occupying state of the object (\coloredtexttt{cup}).
\end{itemize}

\paragraph{Progress Score}
The progress score, denoted as a percentage, reflects the agent's completion rate of required actions within the environment, where the total duration for all actions is considered as 100\%. 
Each action's contribution to the progress score is proportionate to its duration. 
Specifically, if an action's duration is \(t_i\) minutes, its contribution to the progress score is calculated as \(s_i = \left(\frac{t_i}{\sum_{j=1}^{n} t_j}\right) \times 100\%\), with \(n\) representing the total number of actions. 
For instance, an action lasting 5 minutes in a total action duration of 20 minutes contributes 25\% to the progress score.

\section{Experiments}
\label{sec:experiment}
\setlength\tabcolsep{3pt}

\begin{table}[t]
\centering
\small
\begin{tabular}{lccc}
\toprule
\textbf{Scenario} & \textbf{\# Actions} & \textbf{\# Objects} & \textbf{Time (min)}\\
\midrule
Cooking & 5.6 & 5.5 & 18.9\\
Household Activity & 4.1 & 3.5 & 12.8\\
Laboratory Work& 5.3 & 2.7 & 16.1\\
\bottomrule
\end{tabular}
\caption{
Average number of actions and objects per task in each scenario, and the average shortest completion time for these tasks.}
\label{tab:sta} 
\end{table}
\subsection{Experiment Settings}

\paragraph{Task Set Construction}
In our experiments, we design three categories of task combinations based on the number of tasks: \textbf{\# Task=1}, \textbf{\# Task=2} and \textbf{\# Task=3} scenarios.
In \# Task=1 scenario, agents focus on completing one task (\eg, \coloredtexttt{make tea}).
For the other two scenarios, we combine either two or three tasks from 10 single tasks (\eg, \coloredtexttt{make tea} and \coloredtexttt{wash clothes}).
Then, we randomly select 10 combined tasks for each scenario.\footnote{Appendix~\ref{app:example_task} shows examples of single and combined tasks.}

\paragraph{Interaction}
Initially, the environment provides a comprehensive task instruction that details the task, action space, and object set. 
Subsequently, the agent produces an action based on this instruction, adhering to a prescribed format specified in the action space; any deviation is considered invalid. 
To facilitate action recognition by the environment, regular expressions are employed to parse actions from responses (\eg, extracting \coloredtexttt{wash clothes} from \textit{``I will wash clothes''}). 
For each action execution, the agent must incorporate task instructions, previous actions, and feedback from the environment into LLMs as context.\footnote{Appendix~\ref{app:example_inter} gives an example of interaction between the agent and the environment.}

\paragraph{Maximum Time}
Each combined task is allocated a maximum completion time.
We set the time limit for completing a single task at 40 minutes, which exceeds the total time required for all actions in any given task. 
For tasks that are combined, the time limit is proportionally increased by the number of tasks involved.

\paragraph{Oracle Performance}
As shown in Tabel~\ref{tab:main}, \textit{Orcale} represents the optimal performance, including the shortest completion time and the fastest completion rate, which are manually calculated.
Specifically, we calculate oracle performance using a greedy strategy: always start the longest non-occupied actions as early as possible and avoid idleness when there are actions to perform.\footnote{Appendix~\ref{app:algo} shows our algorithm for calculating the oracle performance.}

\newcolumntype{B}{>{\columncolor{blue!4}}c}
\newcolumntype{d}{>{\columncolor{brown!4}}c}
\newcolumntype{q}{>{\columncolor{green!4}}c}

\setlength\tabcolsep{5.7pt}

\begin{table*}[t]
\centering
\small
\begin{tabular}{clBBBBddddqqqq}
\toprule
& \multirow{2}{*}{\textbf{Model}}& \multicolumn{4}{c}{\textbf{\# Task=1}}  & \multicolumn{4}{c}{\textbf{\# Task=2}} & \multicolumn{4}{c}{\textbf{\# Task=3}} \\
\cmidrule(lr){3-6} \cmidrule(lr){7-10} \cmidrule(lr){11-14}
& & \multicolumn{1}{c}{\textbf{AS} $\uparrow$}  & \multicolumn{1}{c}{\cellcolor{white}\textbf{CS} $\uparrow$}&   \multicolumn{1}{c}{\textbf{CR} $\uparrow$} &  \multicolumn{1}{c}{\textbf{CT} $\downarrow$} &  \multicolumn{1}{c}{\textbf{AS} $\uparrow$ }&  \multicolumn{1}{c}{\textbf{CS} $\uparrow$ }&  \multicolumn{1}{c}{\textbf{CR} $\uparrow$ }& \multicolumn{1}{c}{\textbf{CT} $\downarrow$} &  \multicolumn{1}{c}{\textbf{AS} $\uparrow$} & \multicolumn{1}{c}{\textbf{CS} $\uparrow$ }&  \multicolumn{1}{c}{\textbf{CR} $\uparrow$} & \multicolumn{1}{c}{\textbf{CT} $\downarrow$} \\

\midrule

\multirow{10}{*}{\rotatebox{90}{\textbf{Cooking}}}

& Mistral-7B & 63.70 & 3.59 & 30.00 & 25.67 & 42.20 & 1.49 & 0 & - & 39.40 & 1.06 & 0 & - \\ 
& OpenChat-3.5 & 76.30 & \uline{3.89} & 30.00 & \uline{20.33}  & 37.10 & 1.80 & 0 & - & \uline{41.00} & 1.17 & 0 & -  \\
& Vicuna-13B & 84.60 & \textbf{4.10} & \uline{60.00} & 21.83  & 48.80 & 1.76 & 0 & - & 26.00 & 1.03 & 0 & - \\
& Mixtral-8x7B & 50.80 & 3.81 & 10.00 & \textbf{19.00} & 40.10 & \uline{1.99} & 0 & - & 27.60 & 1.17 & 0 & -  \\
\cdashlinelr{2-14}
& Gemini Pro & 78.30 & 3.57 & 50.00 & 24.60 & 31.00 & 1.75 & 0 & - & 18.50 & \textbf{1.26} & 0 & - \\
& GPT-3.5 & 77.70 & 3.61 & 30.00 & 24.33 & 52.30 & 1.87 & 0 & - & 33.10 & \uline{1.23} & 0 & - \\
& GPT-4 & \textbf{98.70} & 3.48 & \textbf{90.00} & 28.22  & \textbf{93.50} & 1.83 & \textbf{70.00} & \uline{52.57}  & \textbf{82.50} & 1.21 & \textbf{40.00} & \textbf{76.25}  \\
& \ \ + Self-plan & \uline{89.00} & 3.83 & \uline{60.00} & 26.50  & \uline{64.90} & \textbf{2.05} & \uline{10.00} & \textbf{37.00}  & 26.20 & 1.15 & 0 & - \\
\cdashlinelr{2-14}
& Oracle & 100 & 5.31 & 100 & 18.90 & 100 & 2.85 & 100 & 35.00 & 100 & 1.94 & 100 & 52.50 \\

\midrule

\multirow{10}{*}{\rotatebox{90}{\textbf{Household Activity}}}
& Mistral-7B & 64.80 & 6.00 & 20.00 & 15.50 & 45.30 & 2.46 & 0 & - & 49.90 &  1.78 & 0 & - \\ 
& OpenChat-3.5 & 70.50 & 5.34 & 30.00 & 15.67 & 68.20 & 2.73 & 0 & - & 44.30 & \uline{1.83} & 0 & - \\
& Vicuna-13B & 69.50 & 5.94 & 40.00 & \textbf{14.25} & 45.90 & 2.34 & 0 & - & 24.90 & 1.69 & 0 & - \\
& Mixtral-8x7B & 68.80 & \textbf{6.08} & 40.00 & \uline{15.00} & 51.60 & 2.85 & 10.0 & \uline{31.00}  & 60.20 & \uline{1.83} & 10.00 & 58.00 \\
\cdashlinelr{2-14}
& Gemini Pro & 68.10 & 5.92 & 40.00 & 16.50 & 60.50 & \textbf{3.02} & 10.00 & \textbf{25.00} & 40.30 & \textbf{1.93} & 0 & - \\
& GPT-3.5 & \uline{87.40} & 5.98 & 70.00 & 16.71 & 63.80 & 2.57 & 10.00 & 36.00  & 45.30 & 1.82 & 0 & - \\
& GPT-4 & \textbf{100} & 5.81 & \textbf{100} & 17.20 & \textbf{100} & \uline{2.89} & \textbf{100} & 34.50  & \textbf{98.40} & 1.82 & \textbf{90.00} & \uline{54.78}  \\
& \ \ + Self-plan  & 87.20 & \uline{6.01} & \uline{80.00} & 16.37  & \uline{84.50} & 2.80 & \uline{50.00} & 35.20  & \uline{95.30} & \textbf{1.93} & \uline{60.00} & \textbf{50.16} \\
\cdashlinelr{2-14}
& Oracle & 100 & 7.81 & 100 & 12.80 & 100 & 4.23 & 100 & 23.60 & 100 & 2.82 & 100 & 35.40 \\

\midrule

\multirow{10}{*}{\rotatebox{90}{\textbf{Laboratory Work}}}
& Mistral-7B & 70.80 & 4.39 & 30.00 & 21.67 & 47.10 & 2.27 & 0 & -  & 38.40 & 1.37 & 0 & - \\ 
& OpenChat-3.5 & 65.50 & 5.07 & 30.00 & \textbf{13.33} & 45.80 & 2.10 & 0 & - & 27.50 & 1.30 & 0 & - \\
& Vicuna-13B & 59.60 & 3.94 & 20.00 & 26.00 & 20.80 &1.87 & 0 & - & 22.90 & 1.40 & 0 & - \\
& Mixtral-8x7B & 64.10 & 4.57 & 40.00 & 24.25  & 41.80 & 2.43 & 0 & -  & 32.40 & 1.58 & 0 & - \\
\cdashlinelr{2-14}
& Gemini Pro & 88.00 & \uline{5.17} & 70.00 & 19.57 & 57.50 & \uline{2.64} & \uline{20.00} & \textbf{35.50} & 25.70 & 1.61 & 0 & - \\
& GPT-3.5 & 71.50 & 4.52 & 30.00 & 22.00  & 47.60 & 2.17 & 0 & - & 37.90 & 1.52 & 0 & - \\
& GPT-4 & \textbf{97.50} & \textbf{5.32} & \textbf{90.00} & \uline{18.67}  & \textbf{85.30} & 2.61 & \textbf{50.00} & 39.20 & \textbf{83.10} & \uline{1.71} & \textbf{60.00} & \uline{60.33}  \\
& \ \ + Self-plan & \uline{95.30} & 5.09 & \uline{80.00} & 20.12 & \uline{83.00} & \textbf{2.79} & \textbf{50.00} & \uline{36.40} & \uline{70.00} & \textbf{1.87} & \textbf{60.00} & \textbf{54.66} \\
\cdashlinelr{2-14}
& Oracle & 100 & 6.21 & 100 & 16.1 & 100 & 4.14 & 100 & 24.60 & 100 & 2.84 & 100 & 35.50 \\

\bottomrule
\end{tabular}
\caption{Model performance under different task combination settings in \method. We report Average Progress Score (\textbf{AS}), Completion Speed (\textbf{CS}), Task Completion Rate (\textbf{CR}), and Average Completion Time (\textbf{CT}). \textbf{\#Task=n} represents that agents are required to do $n$ tasks altogether.
We also list the Oracle result for comparison.
The best results are \textbf{bolded}, and the second best ones are \uline{underlined}. 
}
\label{tab:main}
\end{table*}

\paragraph{Finishing}
The interaction finishes under any of the following conditions:
\begin{inparaenum}[1)]
    \item 
    Agents have completed all the actions that solve the tasks
    (\ie, the progress score reaches 100\%)
    \item Time has run out;
    \item Agents who have performed incorrect actions 5 times in a row are considered to fail the task.
\end{inparaenum}

\paragraph{Model Choice}
We select a variety of language models to drive the agent: Mistral-7B~\cite{jiang2023mistral} by MistralAI, OpenChat-3.5~\cite{wang2023openchat} which is fine-tuned from a 7B Mistral model fine-tuned with C-RLFT (\texttt{openchat\_3.5}), Vicuna-13B~\cite{chiang2023vicuna} which is fine-tuned from a 13B LLaMA model~\cite{touvron2023llama} with instructions and user-shared conversations, Mixtral-8x7B~\cite{mistralai_2023_mixtral} by MistralAI, which is a Mixture-of-Expert version of Mistral, Gemini Pro by Google~\cite{team2023gemini}, GPT-3.5~\citep{openai2022chatgpt} by OpenAI (\texttt{gpt-3.5-turbo-1106}) and GPT-4~\citep{openai2023gpt4} by OpenAI (\texttt{gpt-4-1106-preview}).

\begin{figure}[t]
    \centering    
    \includegraphics[width=\linewidth]{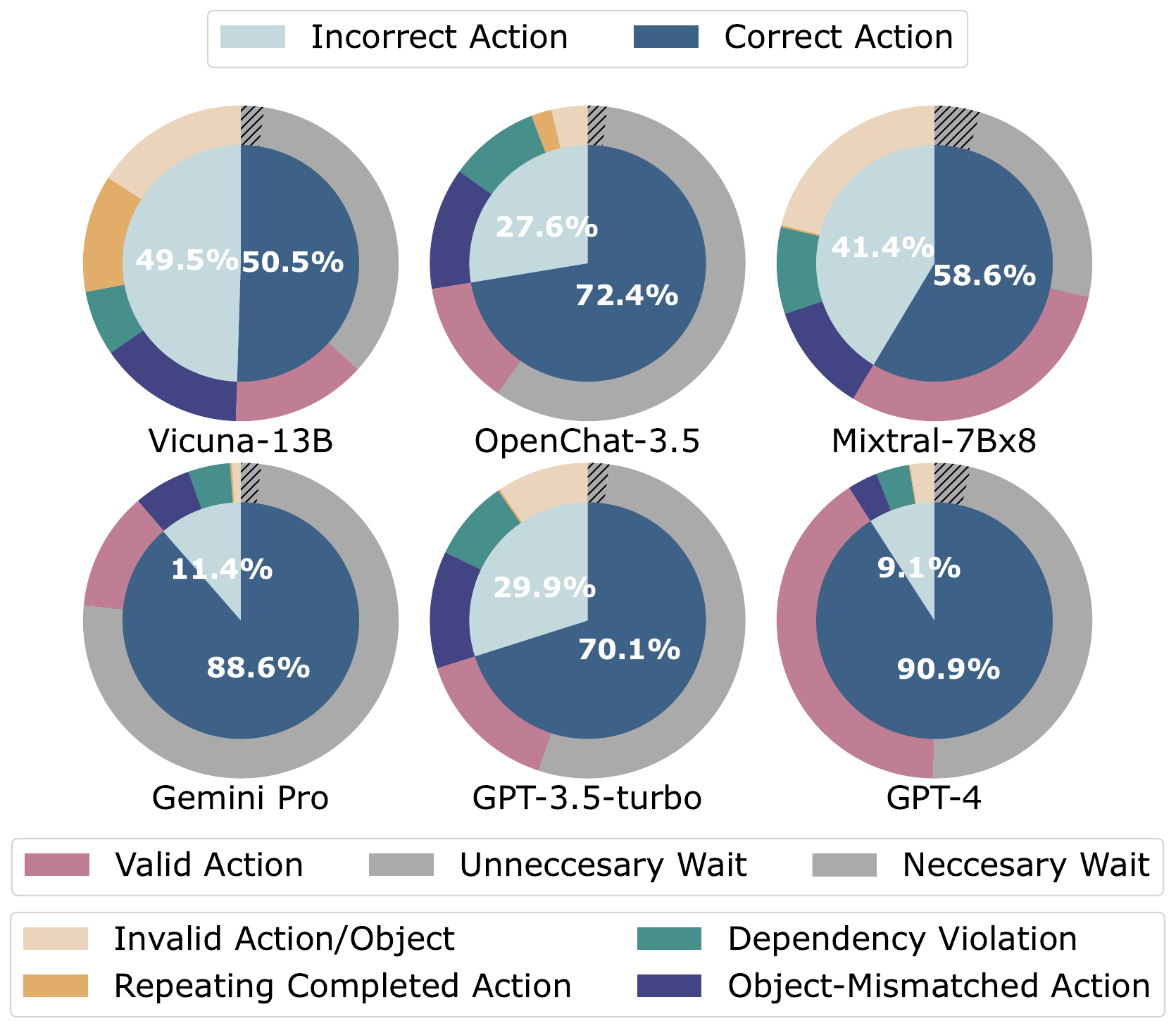}
    \caption{The proportions of correct and incorrect actions of each language agent.
    }
    \label{fig:error}
\end{figure}

\subsection{Evaluation Metrics}
To comprehensively evaluate the ability of agents to multitask, we consider both time and score and design the following four metrics:
\begin{itemize}
    \item \textbf{Average Progress Score (score, AS)}: The average highest progress score achievable by an agent:
            \begin{equation*}
                AS = \frac{\sum_{i \in N} P_i}{N}
            \end{equation*}
        where $P_i$ denotes the maximum progress score of $i$-th task that agents can reach, and $N$ denotes the number of all tasks.  
    \item \textbf{Completion Speed (score per minute, CS)}: The average of the highest score divided by the time taken to achieve it:
        \begin{equation*}
            CS = \frac{\sum_{i \in N} P_i}{\sum_{i \in N} T_i}
        \end{equation*}
    where $T_i$ denotes the time required to reach $P_i$ of $i$-th task.
    \item \textbf{Task Completion Rate (\%, CR)}: The rate of successfully completed tasks:
        \begin{equation*}
            CR = \frac{S}{N}
        \end{equation*}
    where $S$ denotes the tasks completed successfully.
    Notably, when combining tasks, a combined task counts as one task.
    \item \textbf{Average Completion Time (minutes, CT)}: The average time taken for completing tasks successfully:
        \begin{equation*}
            CT = \frac{\sum_{i \in S} T_i}{S}.
        \end{equation*}
\end{itemize}

\subsection{Main Results}

Based on the performance of language agents in Table~\ref{tab:main}, GPT-4 achieves the best performance across different task combinations. 
Moreover, the combined tasks are more challenging than single tasks despite the longer time given.
Apart from GPT-4, most models fail to complete 2 or 3 tasks, showing their limited multitasking abilities and the challenging nature of our environment.

For open-source models, OpenChat-3.5 and Vicuna-13B are even better than GPT-3.5, demonstrating the potential of open-sourced models to develop multitasking capabilities.
However, although OpenChat-3.5 and Vicuna-13B exhibit faster completion speed and shorter average completion time than GPT-4, a lower task completion rate indicates that these models quickly complete simple actions initially but then encounter difficulties;
they either get caught in repetitive actions or fail to properly segment subsequent tasks, which significantly impacts task performance.
For example, initially, \coloredtexttt{potato} is unpicked, so the agent first performs \coloredtexttt{pick potato}. 
Subsequently, the agent mistakenly opts for \coloredtexttt{cook potato in pot} rather than the correct \coloredtexttt{chop potato}, because it incorrectly decomposes the task.

To explore the potential of heuristic algorithms in improving model performance, we introduce \textit{self-plan prompting} to GPT-4, as illustrated in Appendix~\ref{app:example_sp}. 
Under this method, the model initially discovers the dependencies among actions, task descriptions, and objects and estimates the duration of each action. 
It then adopts a greedy strategy similar to \textbf{Oracle Performance}, favoring selecting the longest-duration actions that do not require continuous engagement from the agent in the task model to formulate a plan. Then, the agent executes this plan through interactions with the environment.
However, the results indicate that self-plan prompting is outperformed by vanilla GPT-4.
There are three possible reasons for such performance:
1) The difficulty in accurately parsing actions and identifying their dependencies;
2) The reliance on estimating action durations might introduce cascading errors, leading to inaccurate results of the greedy strategy;
3) The rigid adherence to flawed plans, without adapting to the dynamic nature of interactions with the environment, leads to its failure.

\subsection{Analysis}

\begin{figure}[t]
    \centering
    \includegraphics[width=\linewidth]{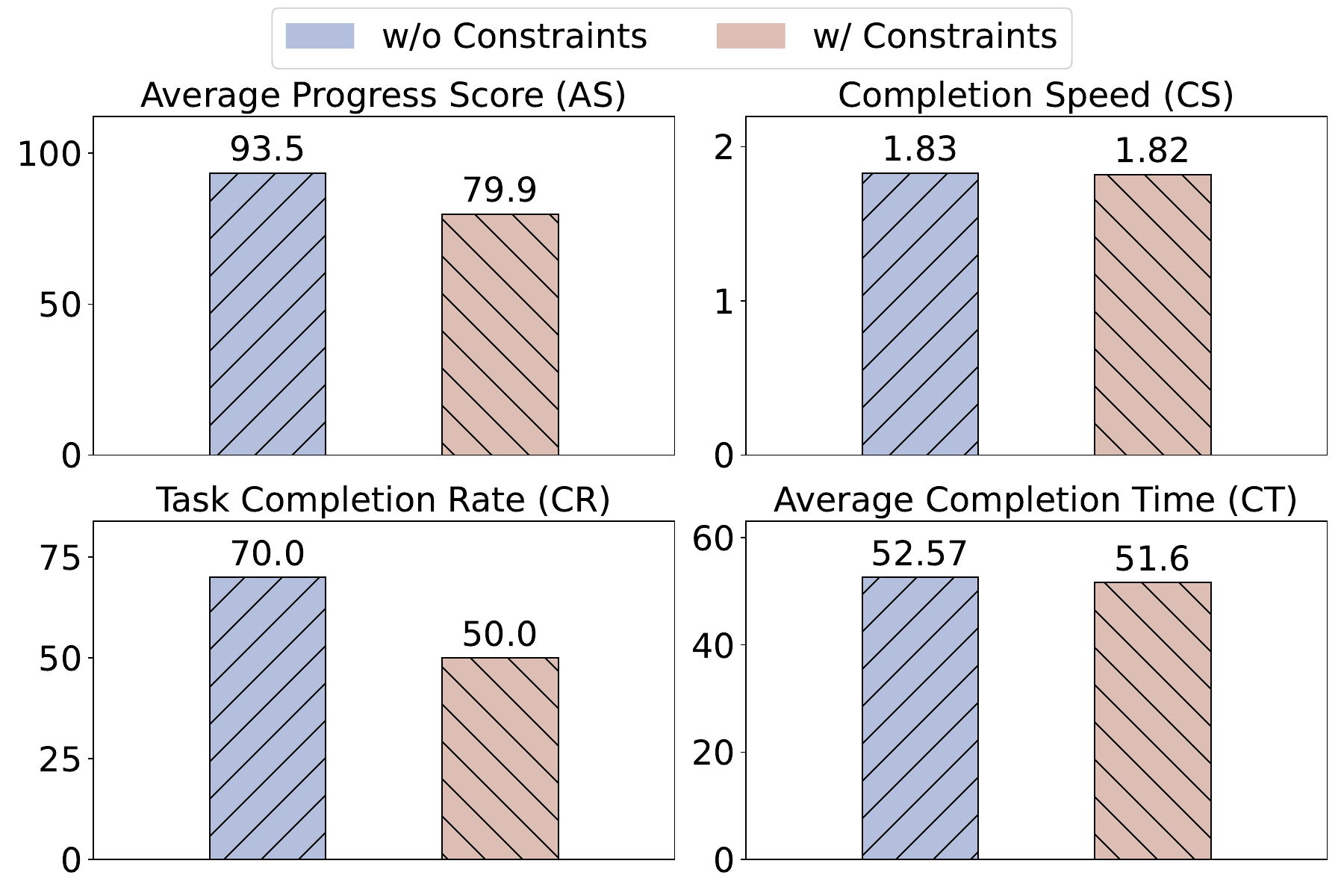}
    \caption{Comparison of the performance of GPT-4 with and without resource constraints. 
    We imposed constraints by limiting to a single instance each of \coloredtexttt{pot}, \coloredtexttt{fryer}, and \coloredtexttt{oven}.
    }
    \label{fig:constraint}
\end{figure}

\begin{figure*}[t]
    \centering
    \includegraphics[width=\linewidth]{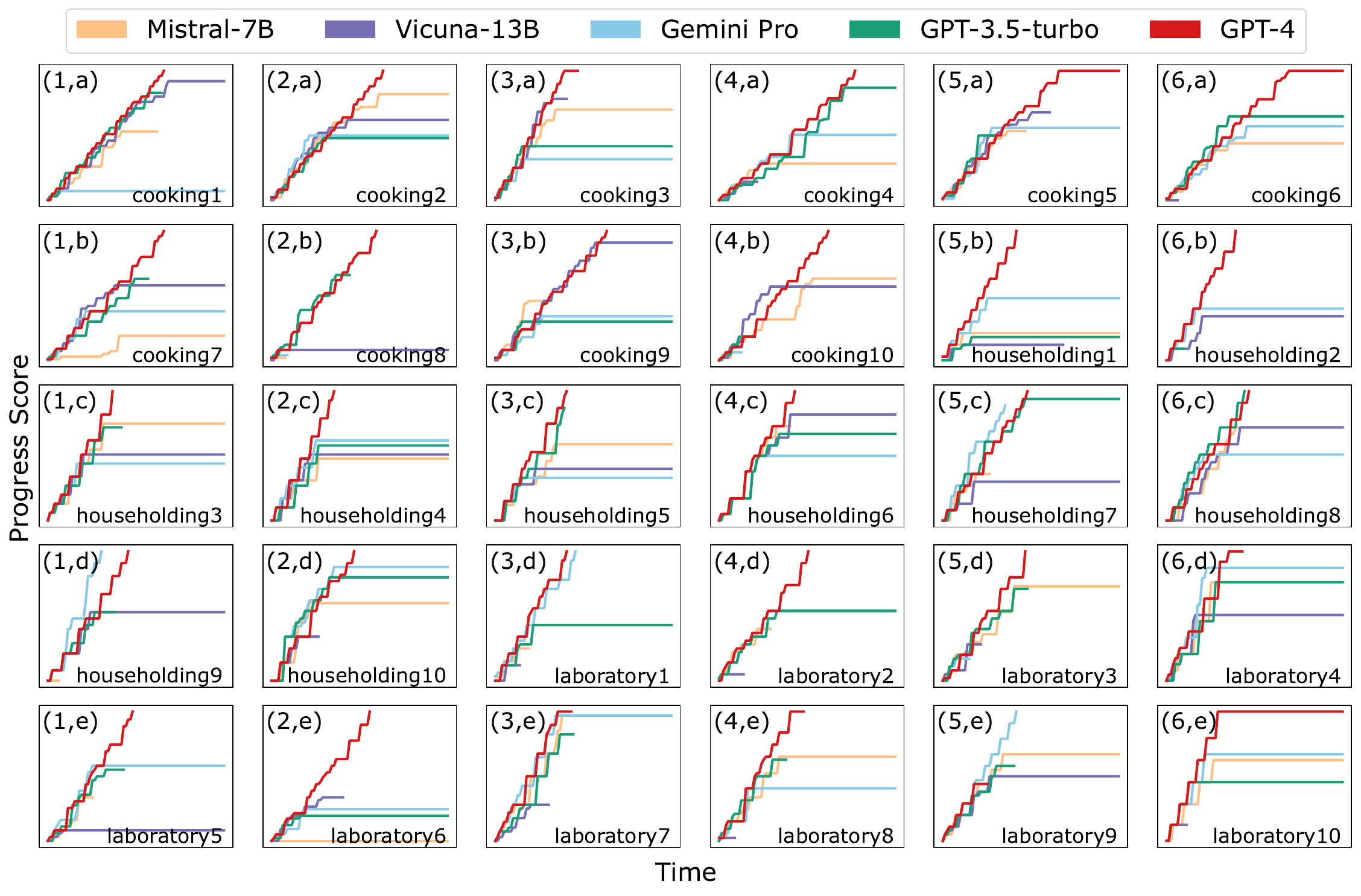}
    \caption{
    Task progress score curves of language agents on two task combinations in \method. 
    The names at the bottom-right indicate the scenario and task number. 
    For example, \texttt{cooking1} represents the first combination of tasks in the cooking scenario.
    }
    \label{fig:progress}
\end{figure*}

\paragraph{Can language agents master multitasking?}
To further explore whether language agents can master multitasking, we conduct detailed analyses to investigate the types of actions these models can perform.
Based on environmental feedback, the actions are classified into two categories: correct ones and incorrect ones.
We further define six fine-grained types of actions:
1) \textbf{Correct Actions}: Valid Action, Wait
2) \textbf{Incorrect Actions}: Invalid Action/Object, Dependency Violation, Repeating Completed Action and Object-Mismacthed Action.\footnote{Detailed description of different types of actions can be found in Appendix~\ref{app:actiontype}.}

We calculate the frequency of these actions of each agent throughout their interactions.
The results in Figure~\ref{fig:error} show that a significant proportion of invalid actions are due to dependency violations and mismatches with objects.
Multitasking involves performing several tasks simultaneously. 
As the number of tasks increases, the complexity of objects and actions escalates, leading to intricate dependencies between actions.
Thus, the high proportion of actions that violate dependencies and mismatch objects suggests that language agents face challenges in managing complex action interdependencies during multitasking, indicating a limitation in their multitasking capabilities.

\paragraph{Are language agents aware of parallel processing?}
Parallel processing can significantly reduce the time required for efficient multitasking. 
As stated in $\mathsection$~\ref{sec3:components}, we consider the duration of time for each action in \method to explore the parallel processing capabilities of language agents.
If an agent is capable of parallel processing, it can engage in additional actions instead of unnecessary waiting for the current action to complete.
To answer this question, we decompose \coloredtexttt{wait} action into two types: \coloredtexttt{necessary wait} and \coloredtexttt{unnecessary wait}.
The former represents that no actions can currently be performed, requiring waiting for other actions to complete. 
In particular, we report the maximum number of \coloredtexttt{necessary wait}. 
\coloredtexttt{Unnecessary wait} indicates that there are other action options available.
Figure~\ref{fig:error} shows that \coloredtexttt{wait} actions constitute over half of the valid actions performed by different LLMs, and \coloredtexttt{necessary wait} only accounts for a small part of it.
This indicates a tendency for agents to engage in unnecessary waiting, showing their ignorance of parallel processing and inability to complete tasks in minimal time (Table~\ref{tab:main}).

\paragraph{Do resource constraints affect the multitasking of language agents?}
Resource constraints refer to limitations in the availability of resources (\eg, the number of objects) necessary for task completion, which is rather common in real life.
To design resource constraints, we first select three objects: \coloredtexttt{pot}, \coloredtexttt{fryer} and \coloredtexttt{oven} in the cooking scenario, and choose \# Task=2 setting in Table~\ref{tab:main}.
Then, we set that there is only one instance of each of the three objects, simulating the limitation of resources in the environment.
Figure~\ref{fig:constraint} compares GPT-4's performance before and after applying these constraints. 
We find that the constraints do not affect the task completion time or speed, revealing that GPT-4 rarely attempts to process tasks in parallel.
However, a noticeable decline in both completion rate and progress score indicates that the constraints prevent the models from better comprehending and decomposing multiple tasks.

\paragraph{Language agents trapped in an infinite loop.}
To delve into why language agents struggle with multiple tasks, we analyze the progress score changes over time for various models.
As illustrated in Figure~\ref{fig:progress}, Vicuna, Mistral, Gemini and GPT-3.5 often cease scoring without completing all tasks, maintaining low scores until time runs out (\eg, (5,b), (2,c) and (6,d)).
We further examine their actions during these periods and find that they always perform \textbf{incorrect actions and waiting} alternately.
Since \coloredtexttt{wait} is a valid action, repeatedly alternating between waiting and incorrect actions does not lead to task failure, but neither does it contribute to an increase in scores.
To find out whether agents wait for good reasons, we ask them to explain each action via the chain-of-thought prompting strategy, and they often believe \coloredtexttt{wait} can pause incorrect actions.
However, they find it hard to adjust their incorrect actions based on feedback after waiting, resulting in them being trapped in infinite loops.

\section{Conclusion}
\label{sec:conclusion}
In this paper, we introduce \method, a text-based simulated environment designed to incorporate the notion of \textit{time}.
\method extends beyond simply acknowledging the dependency of actions by also considering their duration, an essential factor in time modeling.
Using \method, we evaluate the multitasking and parallel processing capability of language agents.
Our findings indicate that as tasks become more complex, the models struggle to complete them and often fail to recognize opportunities for parallel processing. 
This reveals that language agents still have significant room for improvement when completing multiple tasks in dynamic environments, highlighting an area for future research.

\section*{Limitations}
In \method, we implement detailed descriptions of tasks and environments, along with fine-grained textual feedback to simulate interactions.
However, \method is still designed as a textual simulation for LLMs, lacking visual information that might be necessary for agents to succeed in real-world tasks.
For example, in the laboratory work scenario, it is challenging to completely represent chemical reactions through text due to their complexity.
The number of tasks and scenarios is limited, while the number of multitasking scenarios that allow parallel processing is large in real life.
Moreover, in \method, agents interact with the environment only through actions that are explicitly presented in action prompts, rather than exploring freely.
Also, whether an action occupies an agent sometimes depends on specific conditions. 
For instance, the action \coloredtexttt{cook beef} is classified as non-occupying in \method, implying that it does not engage agents continuously. Yet, in reality, this action requires attention, such as turning the beef to prevent burning, a detail \method overlooks, potentially reducing the realism of our simulation.

\section*{Ethical Statement}
We hereby acknowledge that all authors of this work are aware of the provided ACL Code of Ethics and honor the code of conduct.

\paragraph{Use of Human Annotations}
Our institution recruited three annotators to implement the task creation for three scenarios. 
We ensure the privacy rights of the annotators are respected during the annotation process.
The annotators receive compensation exceeding the local minimum wage and have consented to tasks generated for \method for research purposes. 

\paragraph{Risks}
The \method in our experiment is created by human annotators, and we further examine them to guarantee that they are devoid of socially harmful or toxic language. 
However, evaluating the data quality of tasks is based on common sense, which can vary among individuals from diverse backgrounds.

\section*{Acknowledgement}

We would like to thank Xintao Wang, Ruihan Yang, Tinghui Zhu from Fudan University for their valuable comments and suggestions for the manuscript.
We would also like to thank Peter Jansen from University of Arizona and Bodhisattwa Prasad Majumder from Allen Institute for AI for fruitful discussions that helped shape this project at an early stage.

\bibliography{anthology}
\bibstyle{acl_natbib}

\clearpage
\appendix
\section{\method Details}
\subsection{Tasks}
\label{app:task}
\begin{figure*}[t]
    \centering
	\subfigure[The first task in the cooking scenario.] { \label{fig:c1} 
		\includegraphics[width=0.44\linewidth]{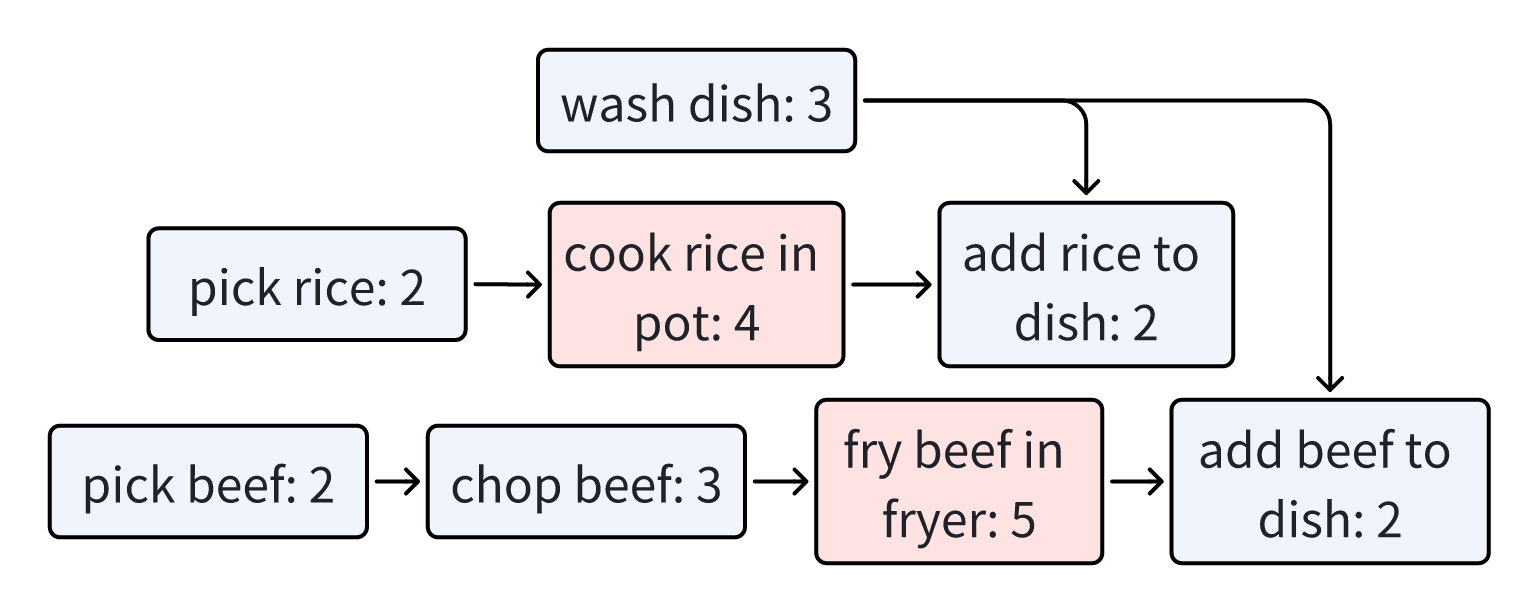}}
	\subfigure[The second task in the cooking scenario.] { \label{fig:c2} 
		\includegraphics[width=0.44\linewidth]{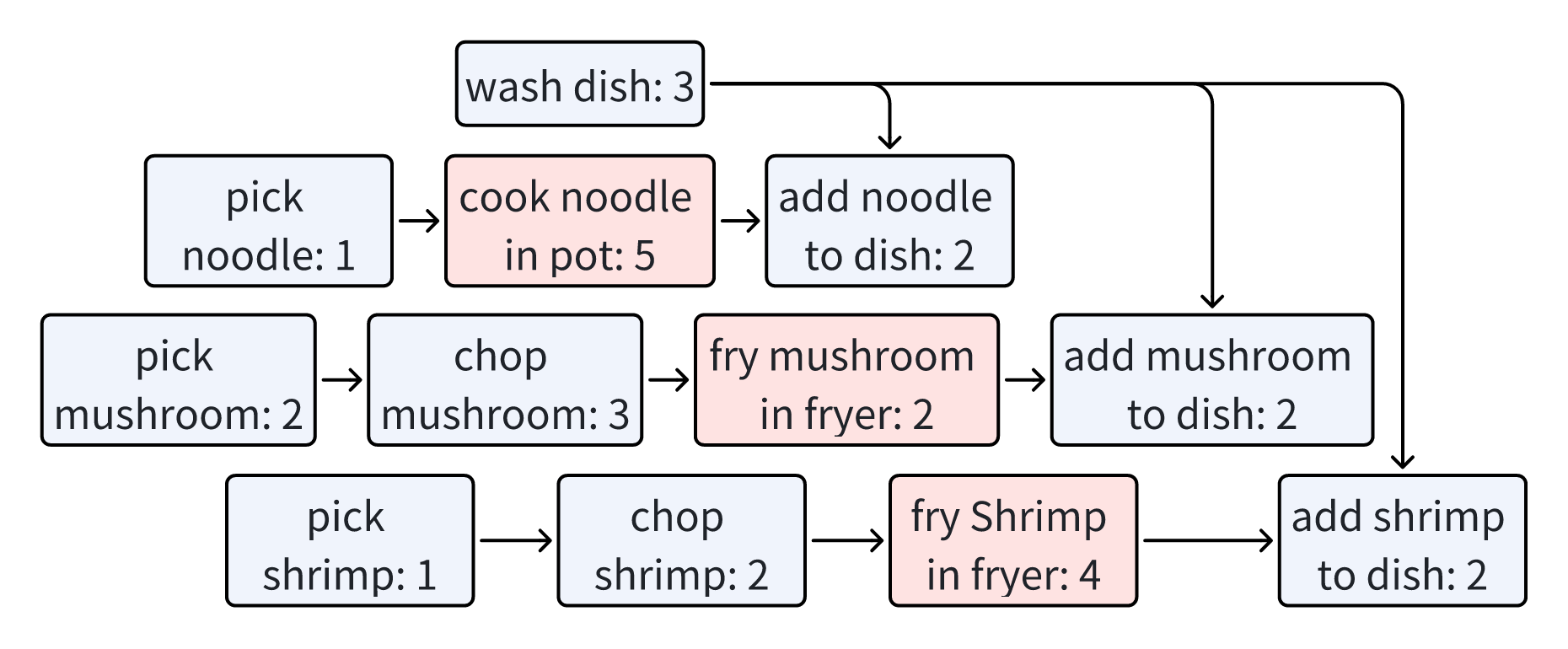}}
        \subfigure[The third task in the cooking scenario.] { \label{fig:c3}
		\includegraphics[width=0.44\linewidth]{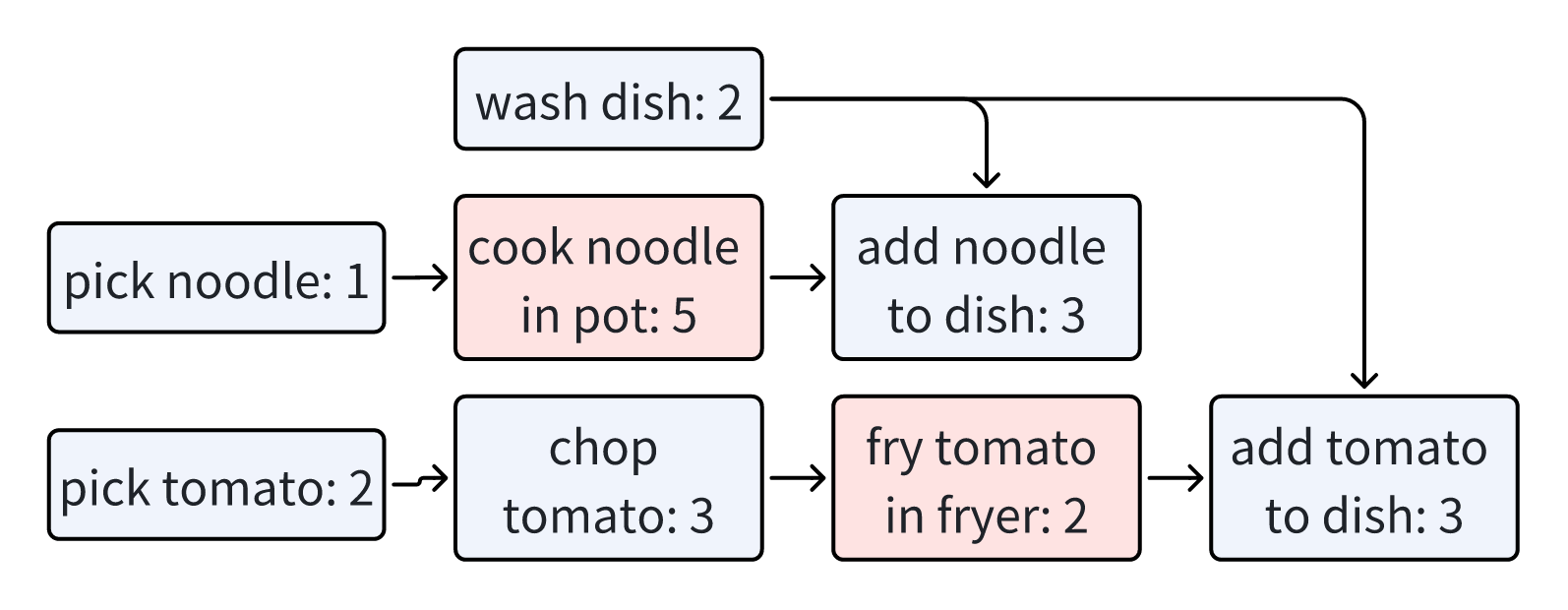}}
        \subfigure[The fourth task in the cooking scenario.] { \label{fig:c4}
		\includegraphics[width=0.44\linewidth]{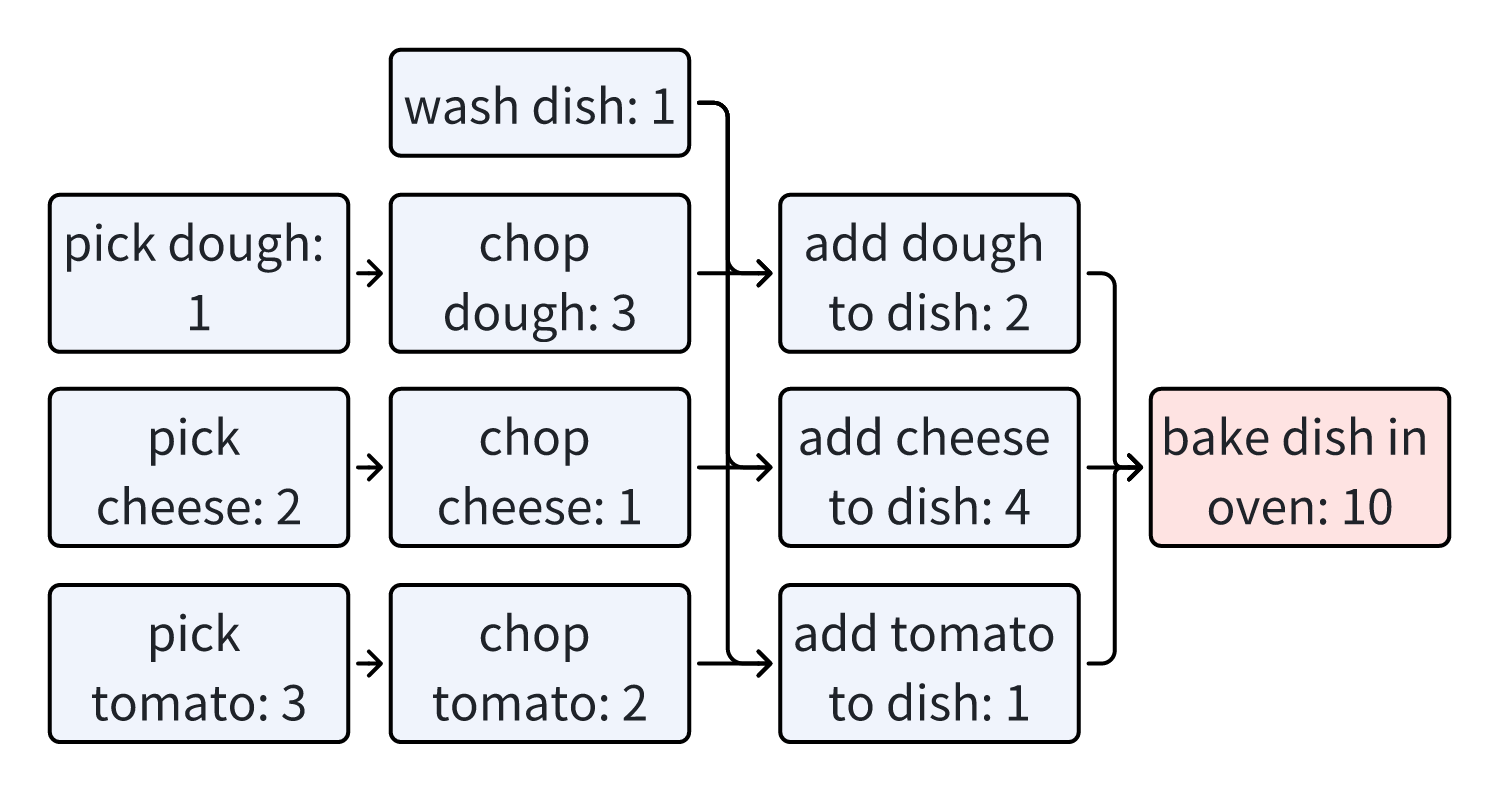}}
        \subfigure[The fifth task in the cooking scenario.] { \label{fig:c5}
		\includegraphics[width=0.44\linewidth]{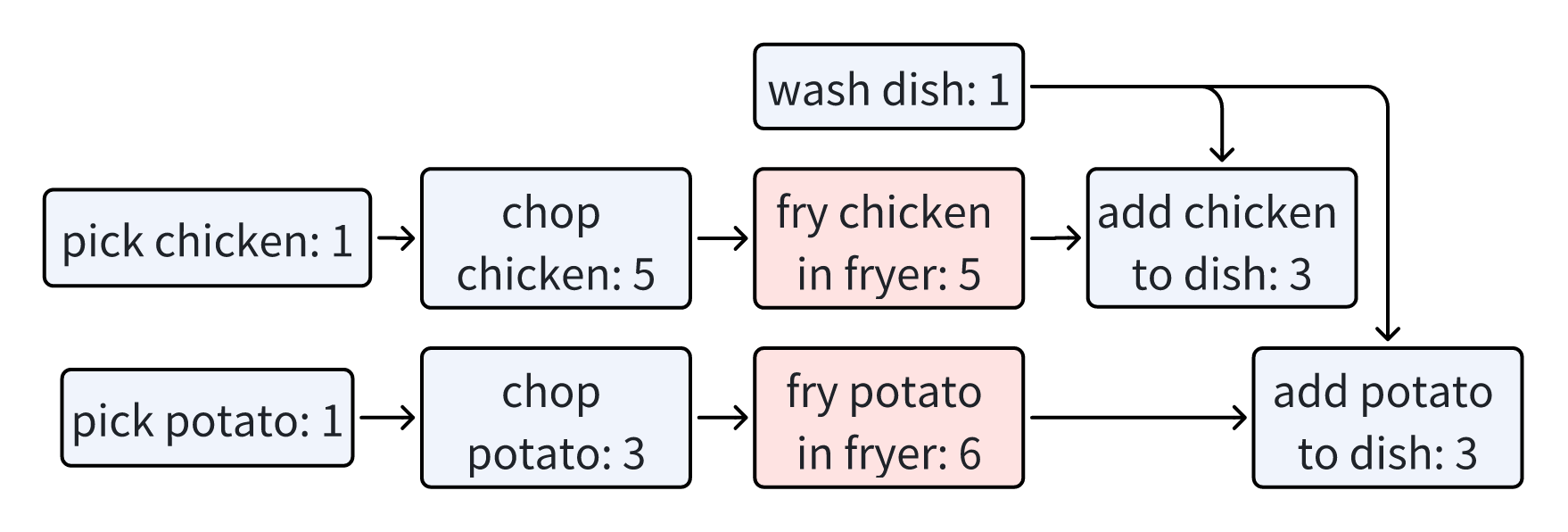}}
        \subfigure[The sixth task in the cooking scenario.] { \label{fig:c6}
		\includegraphics[width=0.44\linewidth]{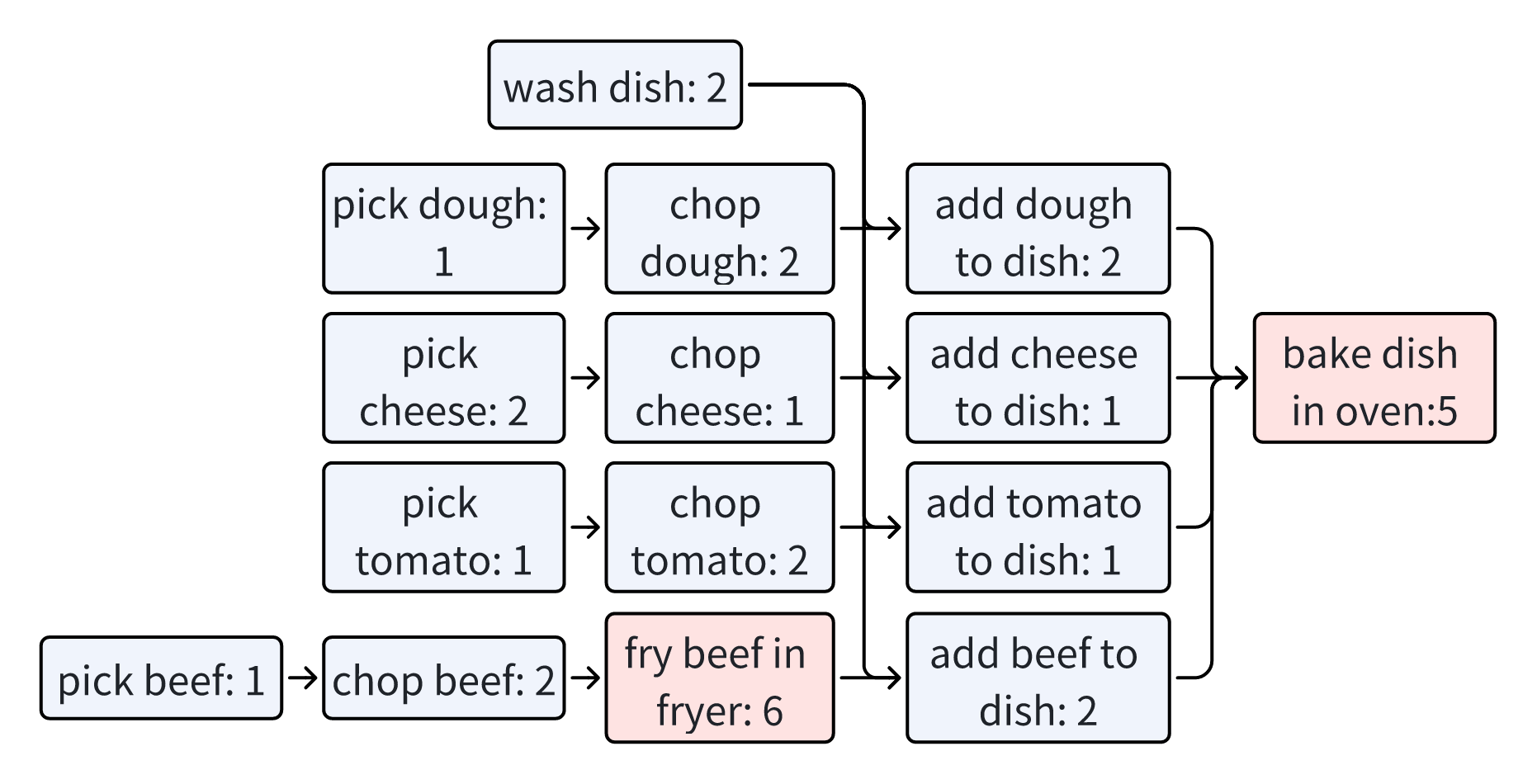}}
        \subfigure[The seventh task in the cooking scenario.] { \label{fig:c7}
		\includegraphics[width=0.44\linewidth]{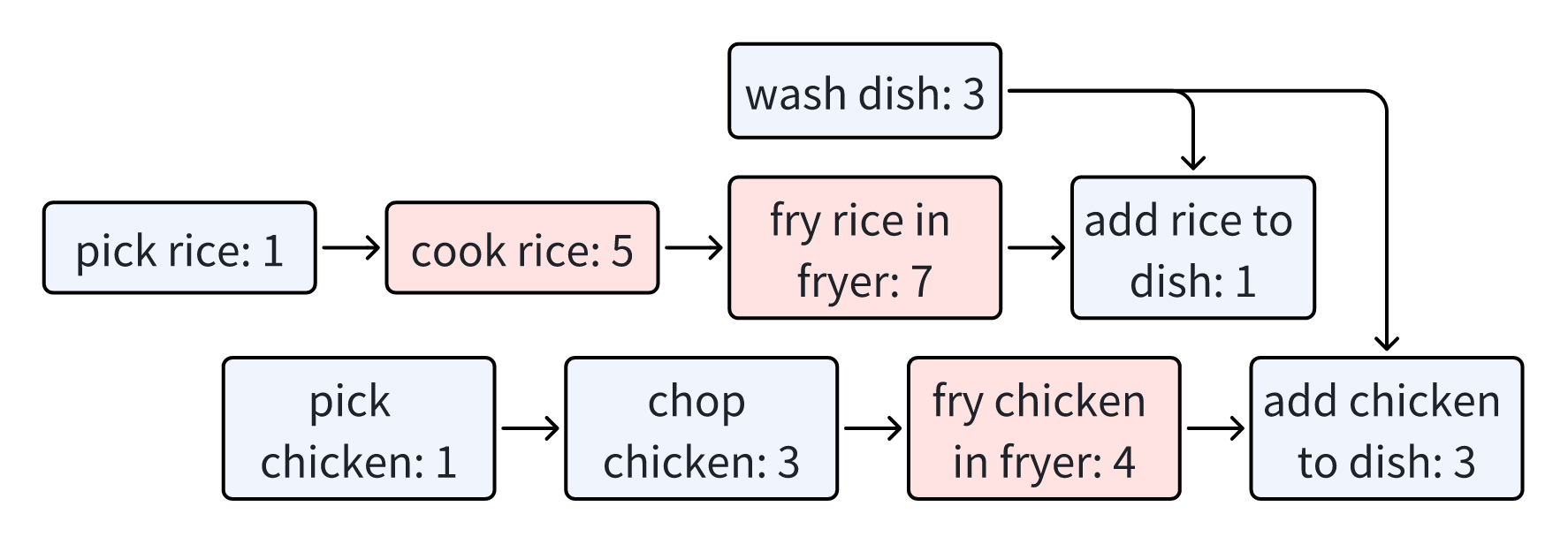}}
        \subfigure[The eighth task in the cooking scenario.] { \label{fig:c8}
		\includegraphics[width=0.44\linewidth]{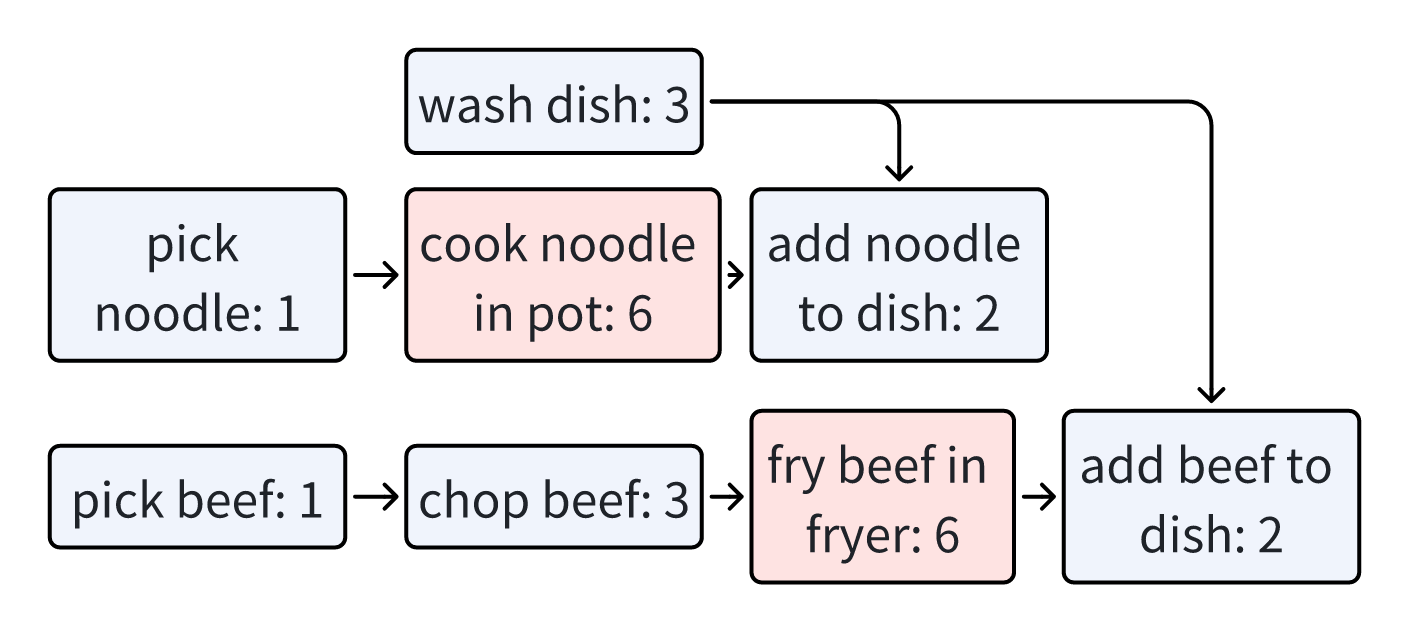}}
        \subfigure[The ninth task in the cooking scenario.] { \label{fig:c9}
		\includegraphics[width=0.44\linewidth]{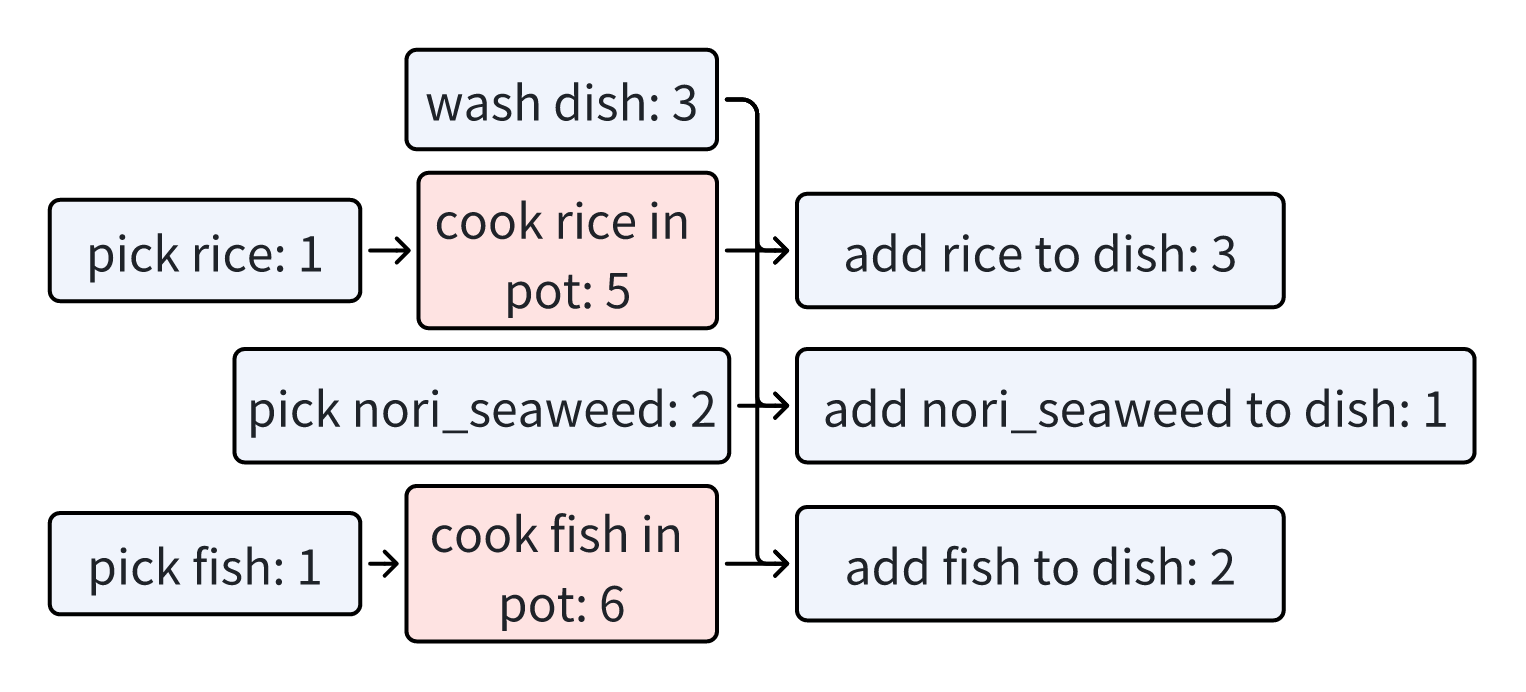}}
        \subfigure[The tenth task in the cooking scenario.] { \label{fig:c10}
		\includegraphics[width=0.44\linewidth]{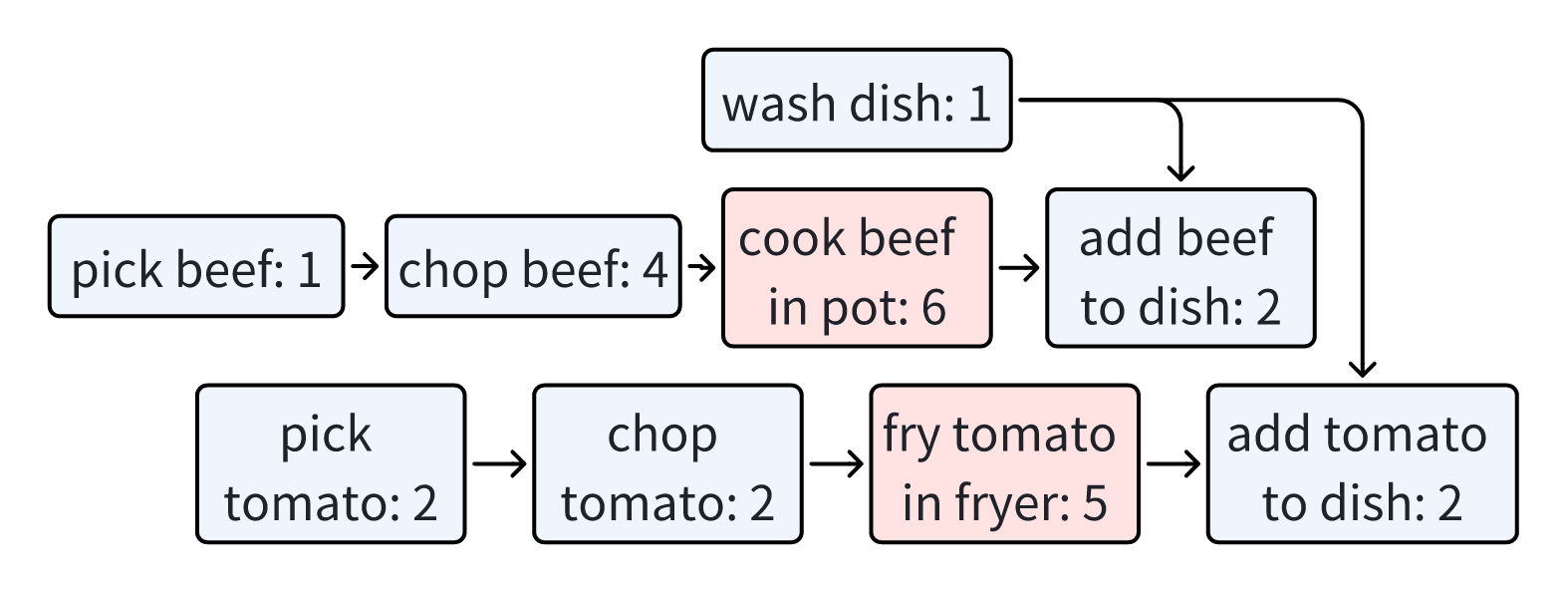}}
    \caption{The action dependencies and durations for the ten tasks in the cooking scenario. Actions that occupy the agent, preventing them from doing anything else, are indicated with a blue background. In contrast, actions not occupying the agent, allowing for parallel tasks, are marked with a red background.}
    \label{fig:cooking_tasks}
\end{figure*}

\begin{figure*}[t]
    \centering
	\subfigure[The first task in the household activity scenario.] { \label{fig:h1} 
		\includegraphics[width=0.44\linewidth]{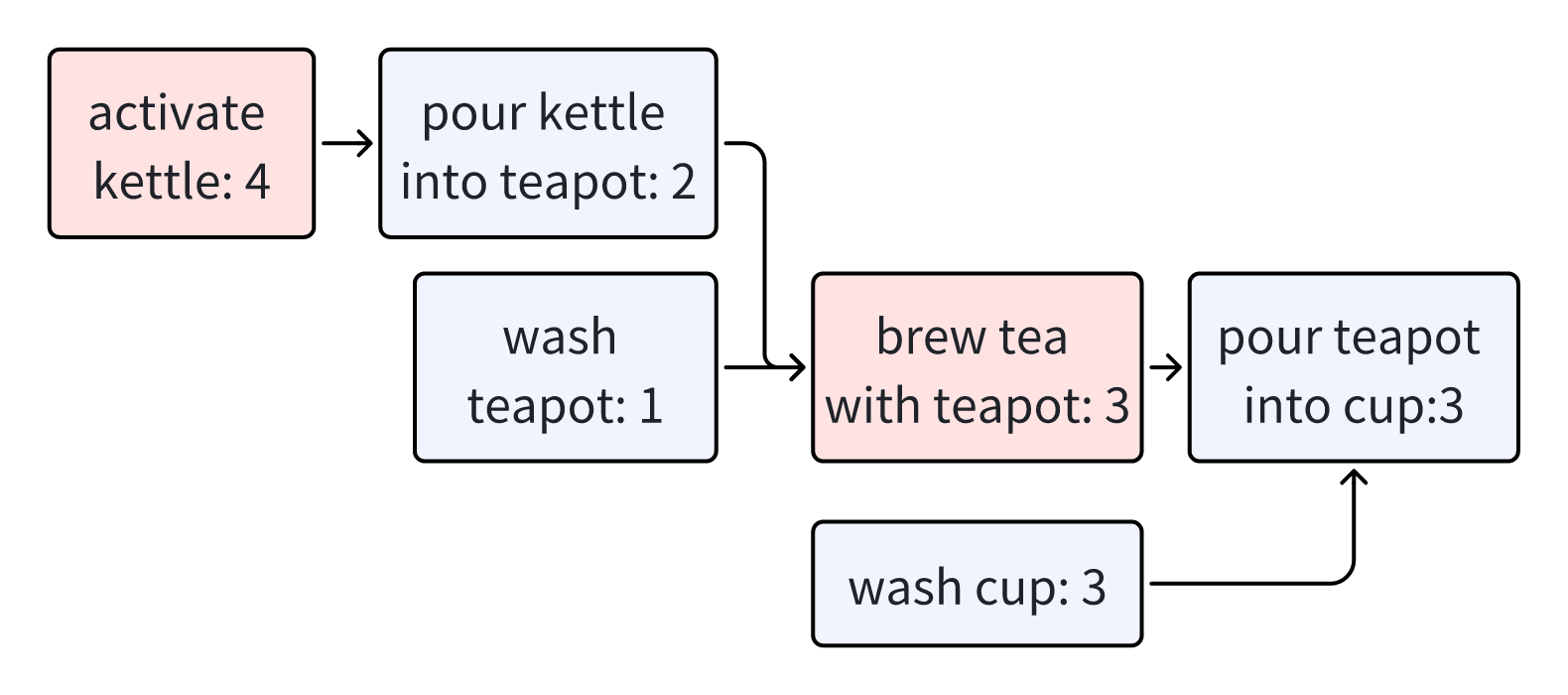}}
	\subfigure[The second task in the household activity scenario.] { \label{fig:h2} 
		\includegraphics[width=0.44\linewidth]{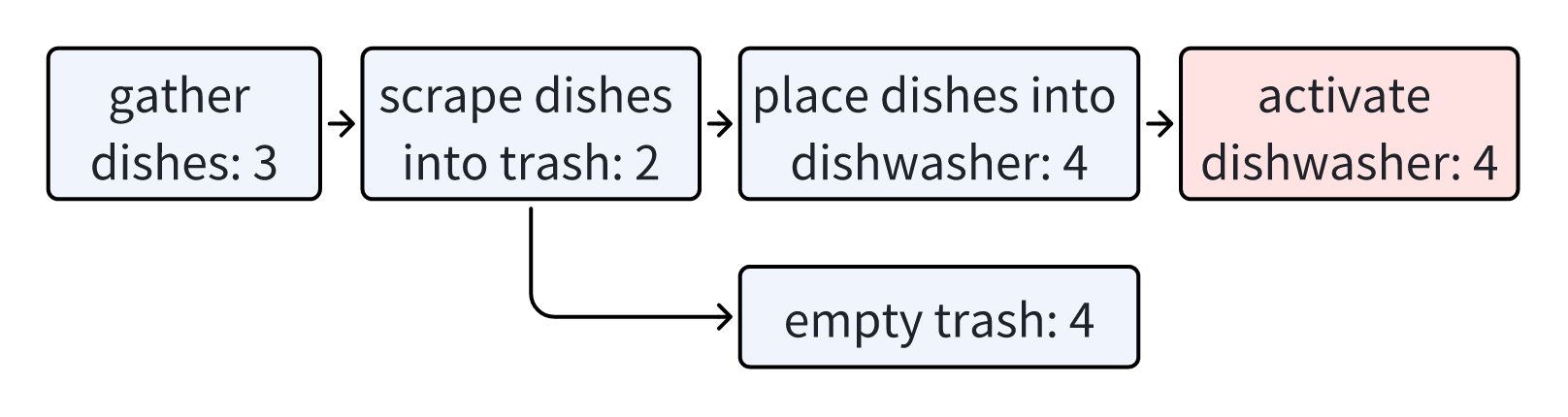}}
        \subfigure[The third task in the household activity scenario.] { \label{fig:h3}
		\includegraphics[width=0.44\linewidth]{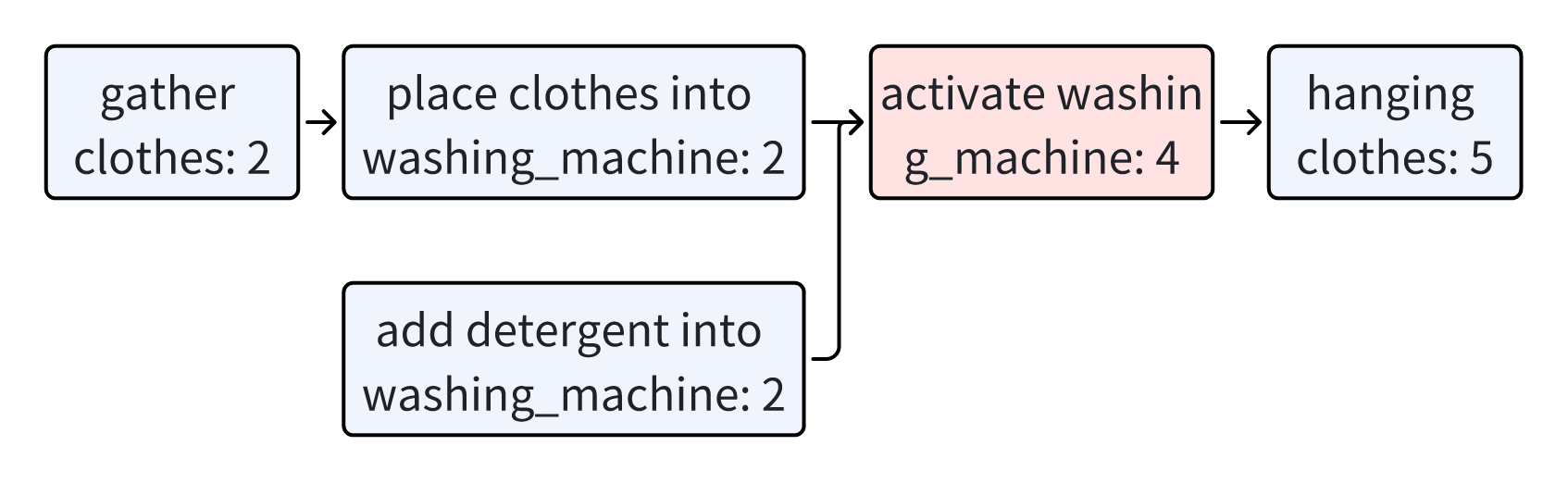}}
        \subfigure[The fourth task in the household activity scenario.] { \label{fig:h4}
		\includegraphics[width=0.44\linewidth]{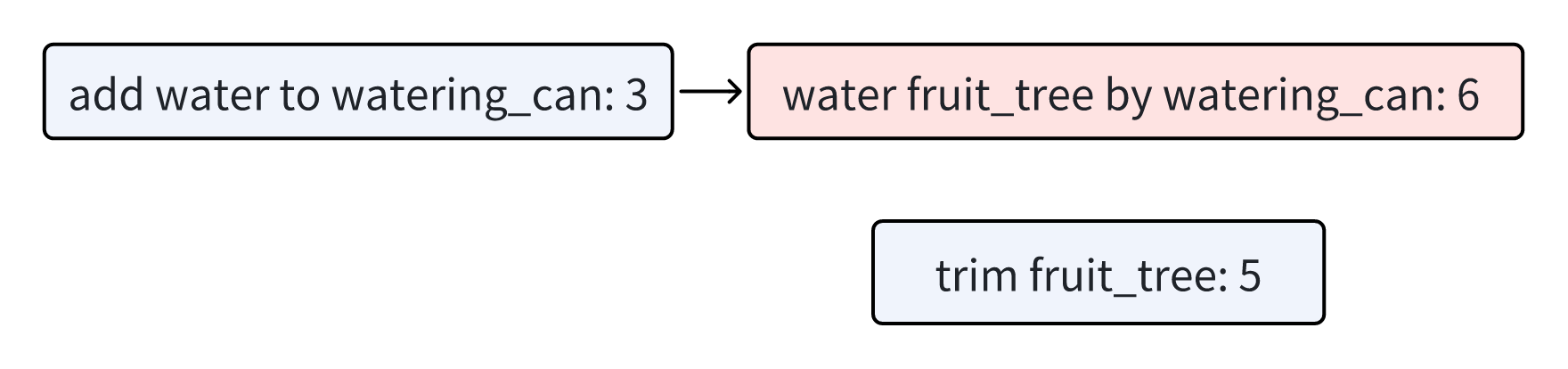}}
        \subfigure[The fifth task in the household activity scenario.] { \label{fig:h5}
		\includegraphics[width=0.44\linewidth]{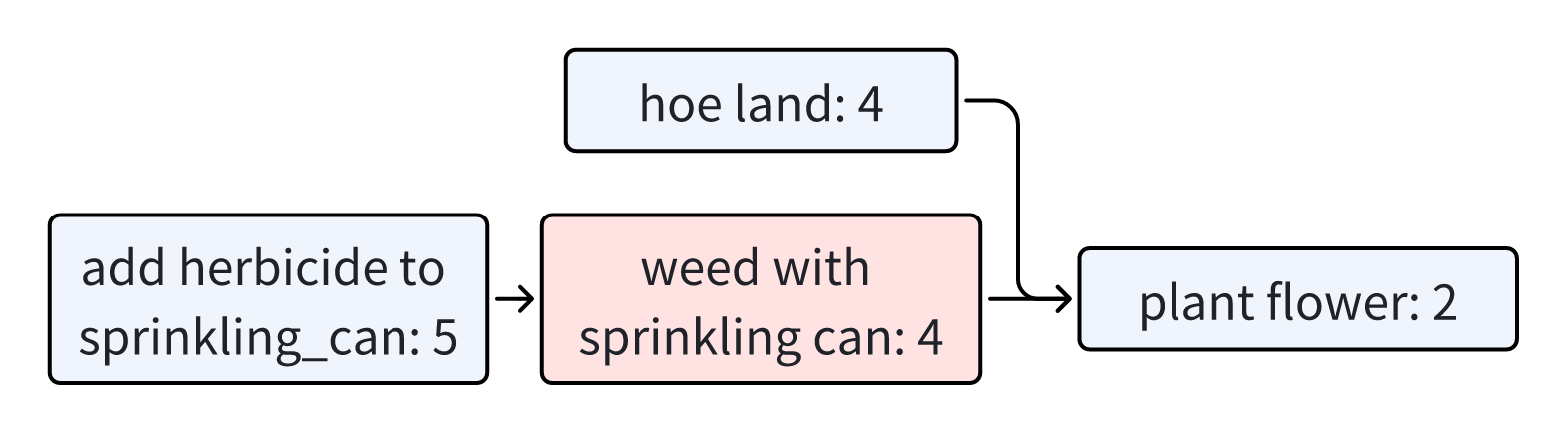}}
        \subfigure[The sixth task in the household activity scenario.] { \label{fig:h6}
		\includegraphics[width=0.44\linewidth]{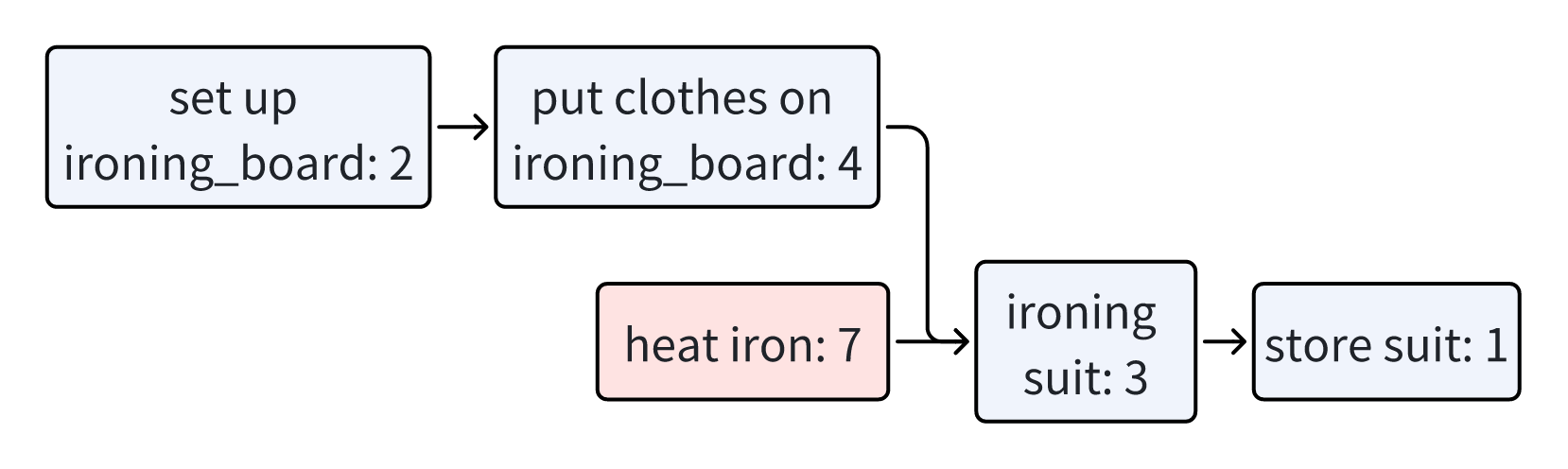}}
        \subfigure[The seventh task in the household activity scenario.] { \label{fig:h7}
		\includegraphics[width=0.44\linewidth]{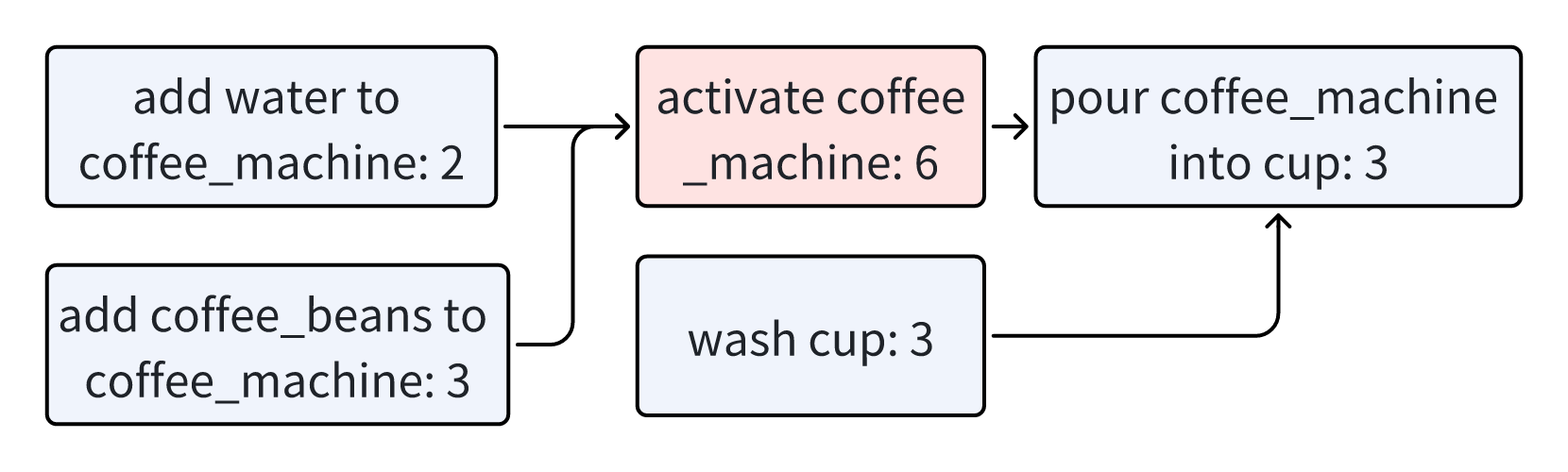}}
        \subfigure[The eighth task in the household activity scenario.] { \label{fig:h8}
		\includegraphics[width=0.44\linewidth]{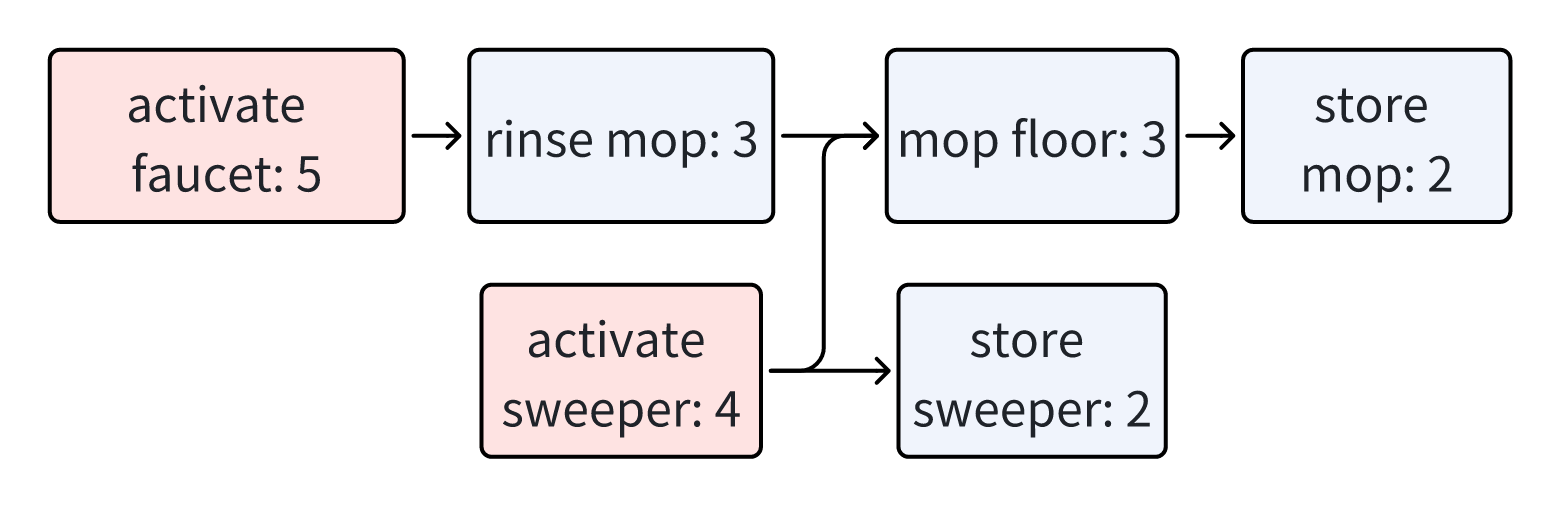}}
        \subfigure[The ninth task in the household activity scenario.] { \label{fig:h9}
		\includegraphics[width=0.44\linewidth]{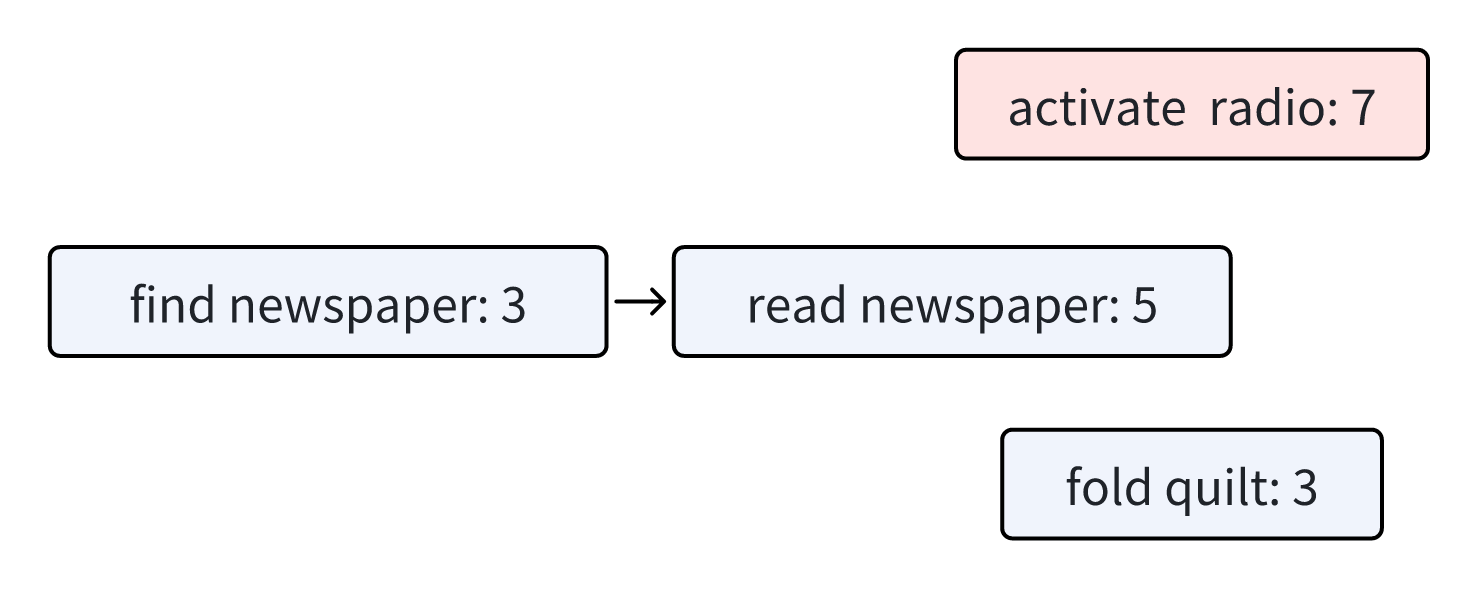}}
        \subfigure[The tenth task in the household activity scenario.] { \label{fig:h10}
		\includegraphics[width=0.44\linewidth]{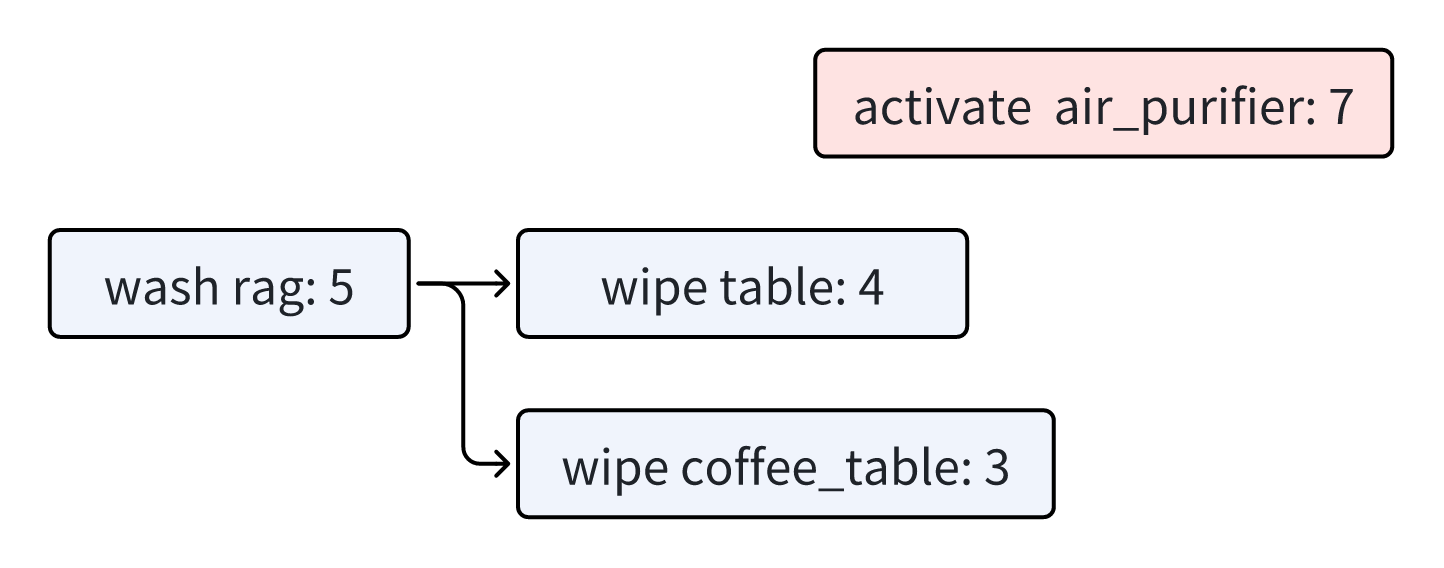}}
    \caption{The action dependencies and durations for the ten tasks in the household activity scenario. Actions that occupy the agent, preventing them from doing anything else, are indicated with a blue background. In contrast, actions that do not occupy the agent, allowing for parallel tasks, are marked with a red background.}
    \label{fig:householding_tasks}
\end{figure*}

\begin{figure*}[t]
    \centering
	\subfigure[The first task in the laboratory work scenario.] { \label{fig:l1} 
		\includegraphics[width=0.44\linewidth]{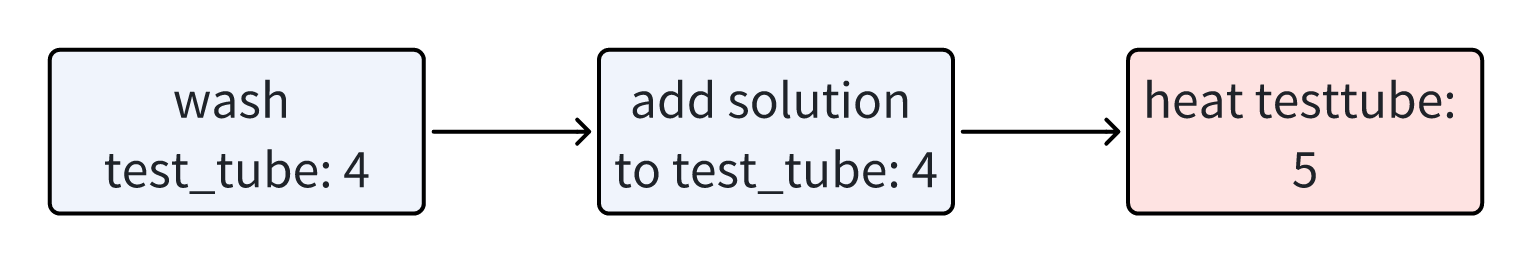}}
	\subfigure[The second task in the laboratory work scenario.] { \label{fig:l2} 
		\includegraphics[width=0.44\linewidth]{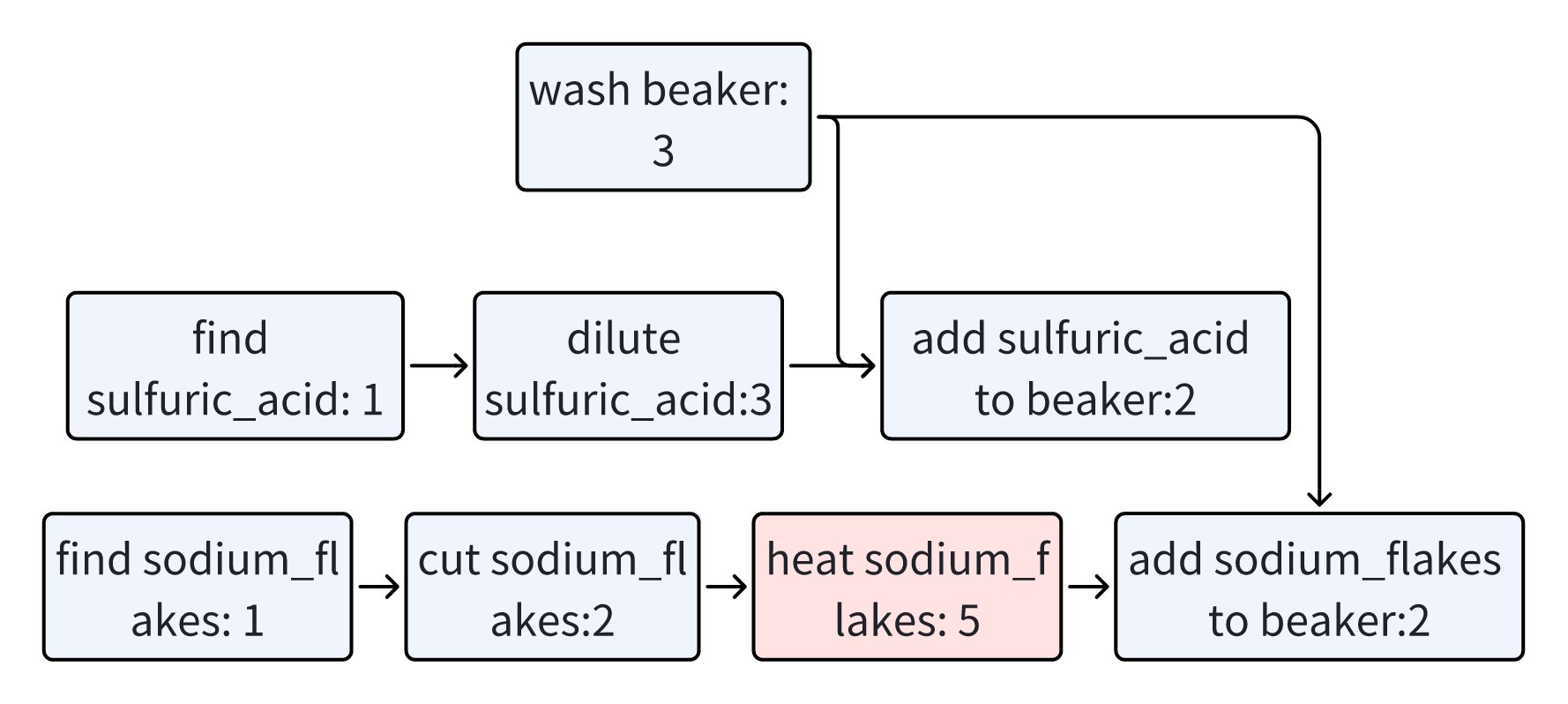}}
        \subfigure[The third task in the laboratory work scenario.] { \label{fig:l3}
		\includegraphics[width=0.44\linewidth]{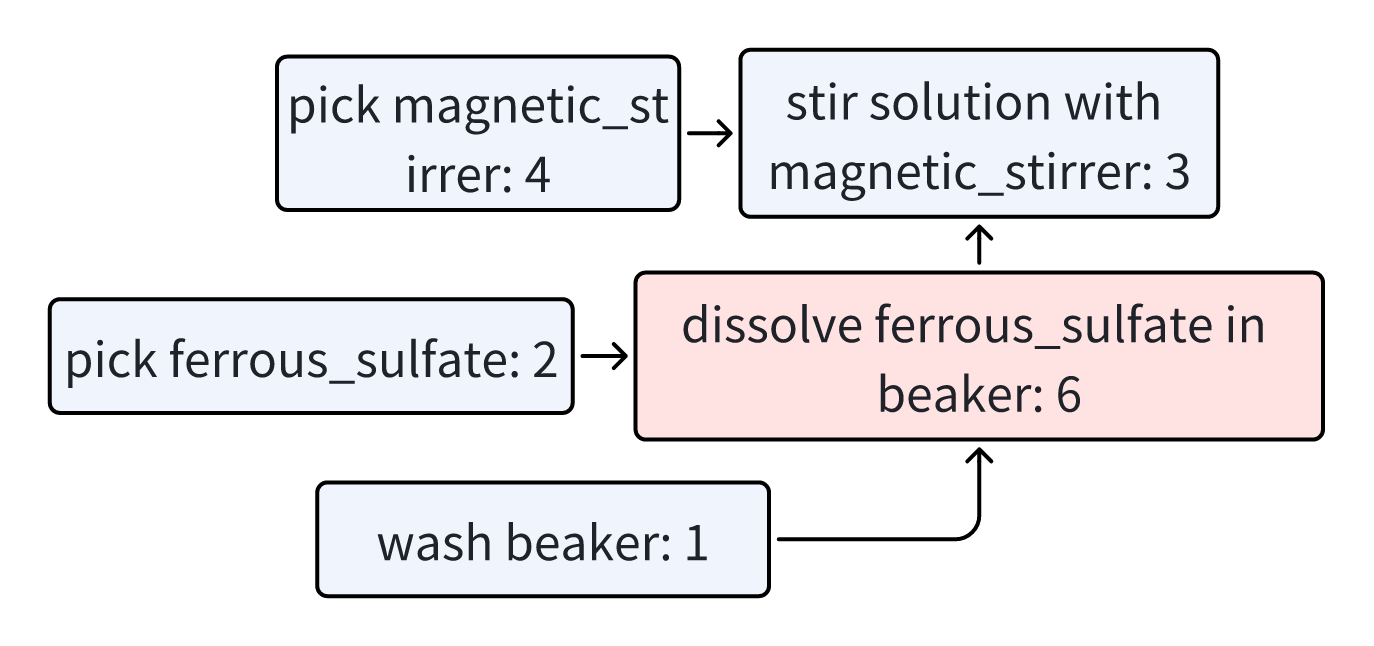}}
        \subfigure[The fourth task in the laboratory work scenario.] { \label{fig:l4}
		\includegraphics[width=0.44\linewidth]{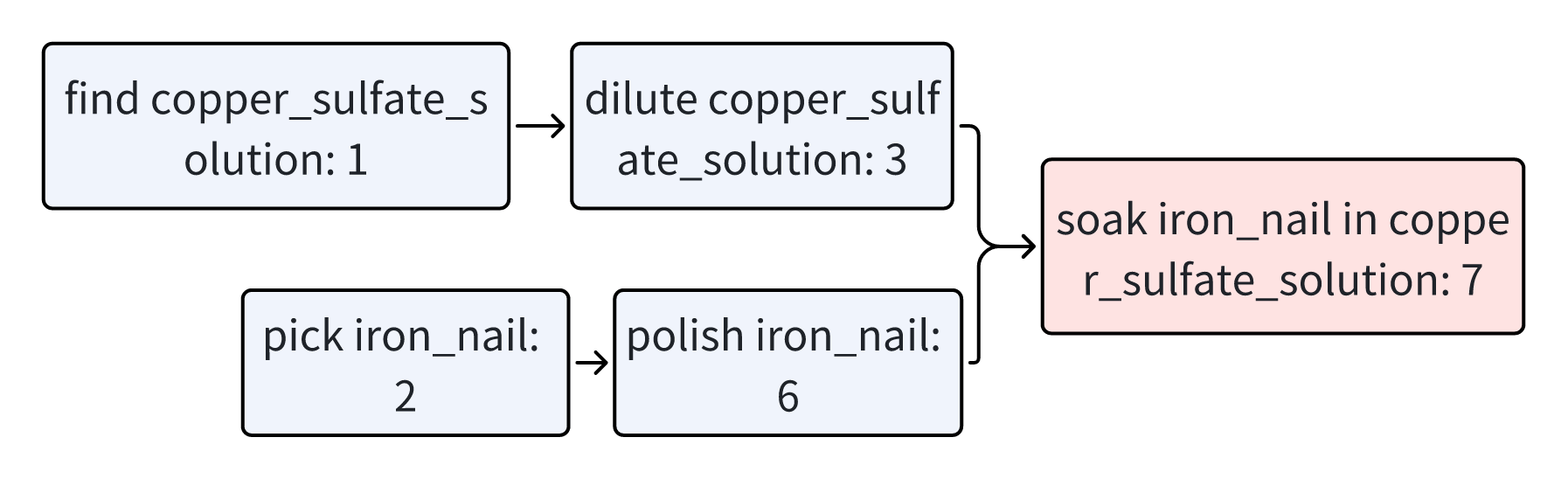}}
        \subfigure[The fifth task in the laboratory work scenario.] { \label{fig:l5}
		\includegraphics[width=0.44\linewidth]{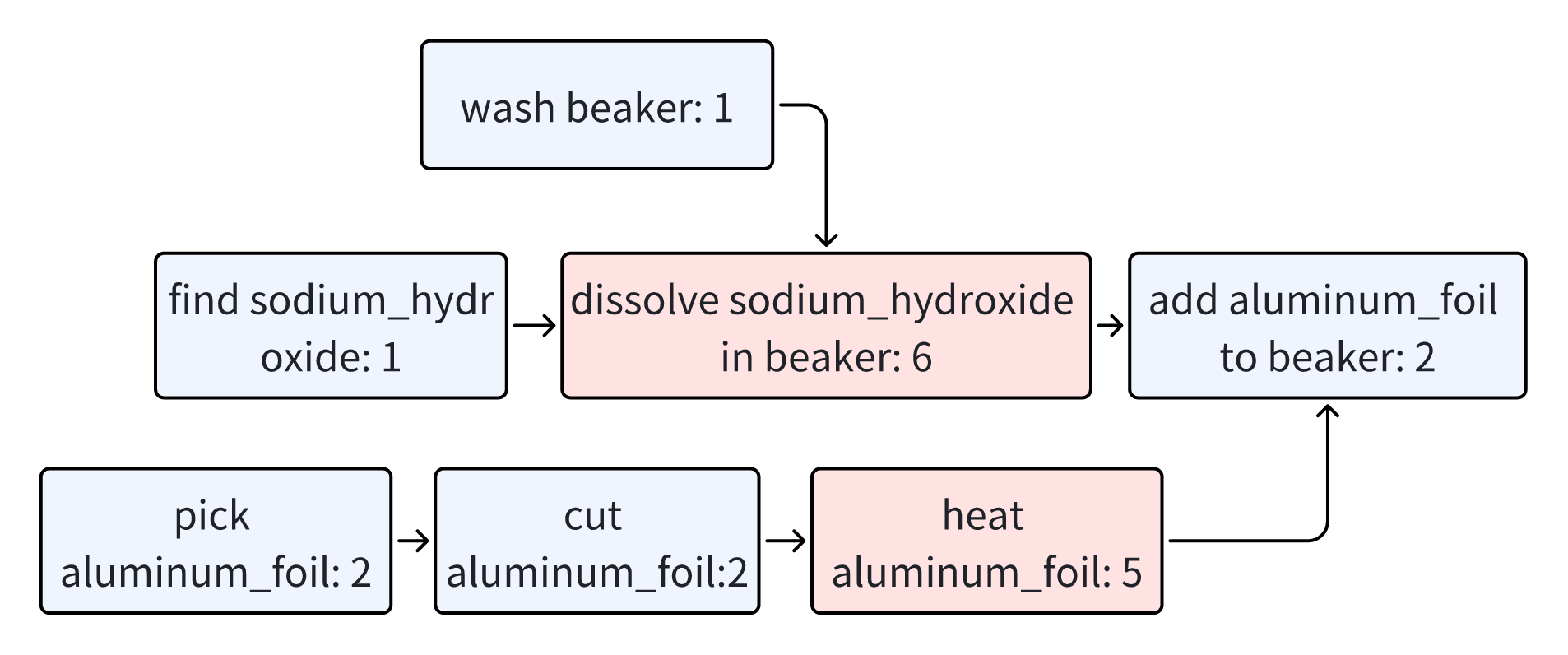}}
        \subfigure[The sixth task in the laboratory work scenario.] { \label{fig:l6}
		\includegraphics[width=0.44\linewidth]{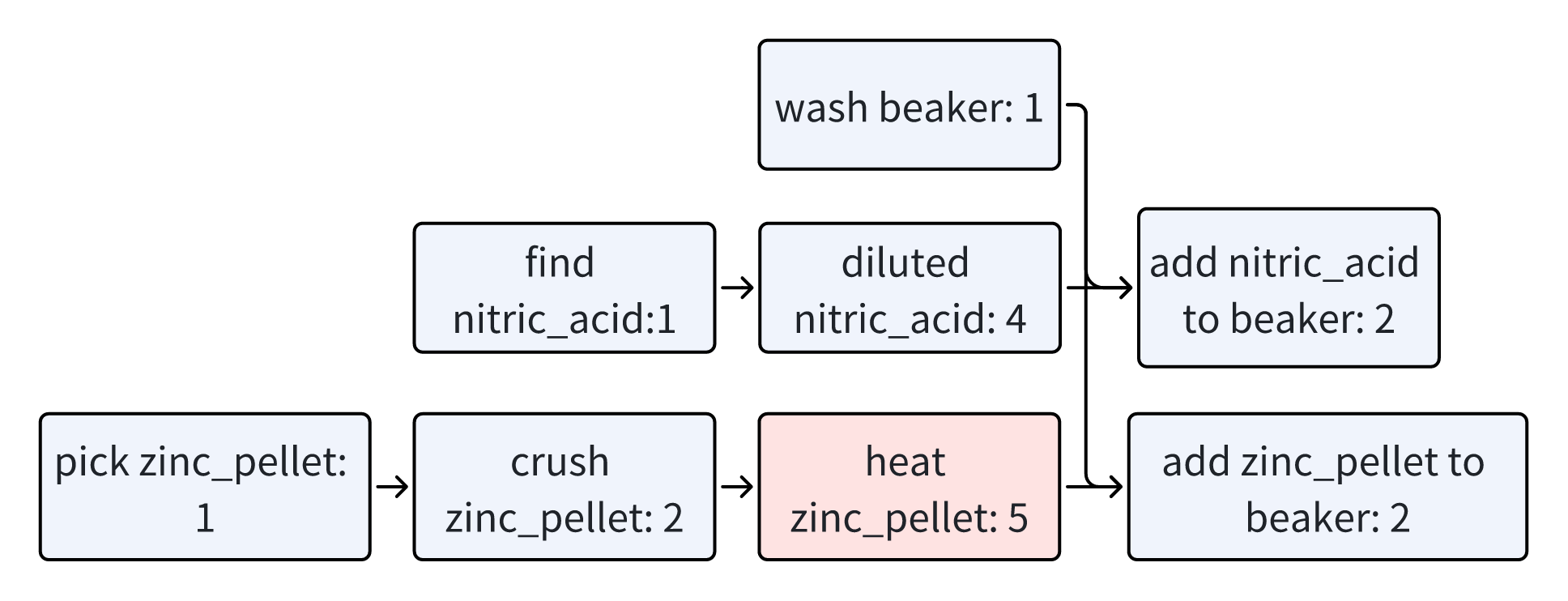}}
        \subfigure[The seventh task in the laboratory work scenario.] { \label{fig:l7}
		\includegraphics[width=0.44\linewidth]{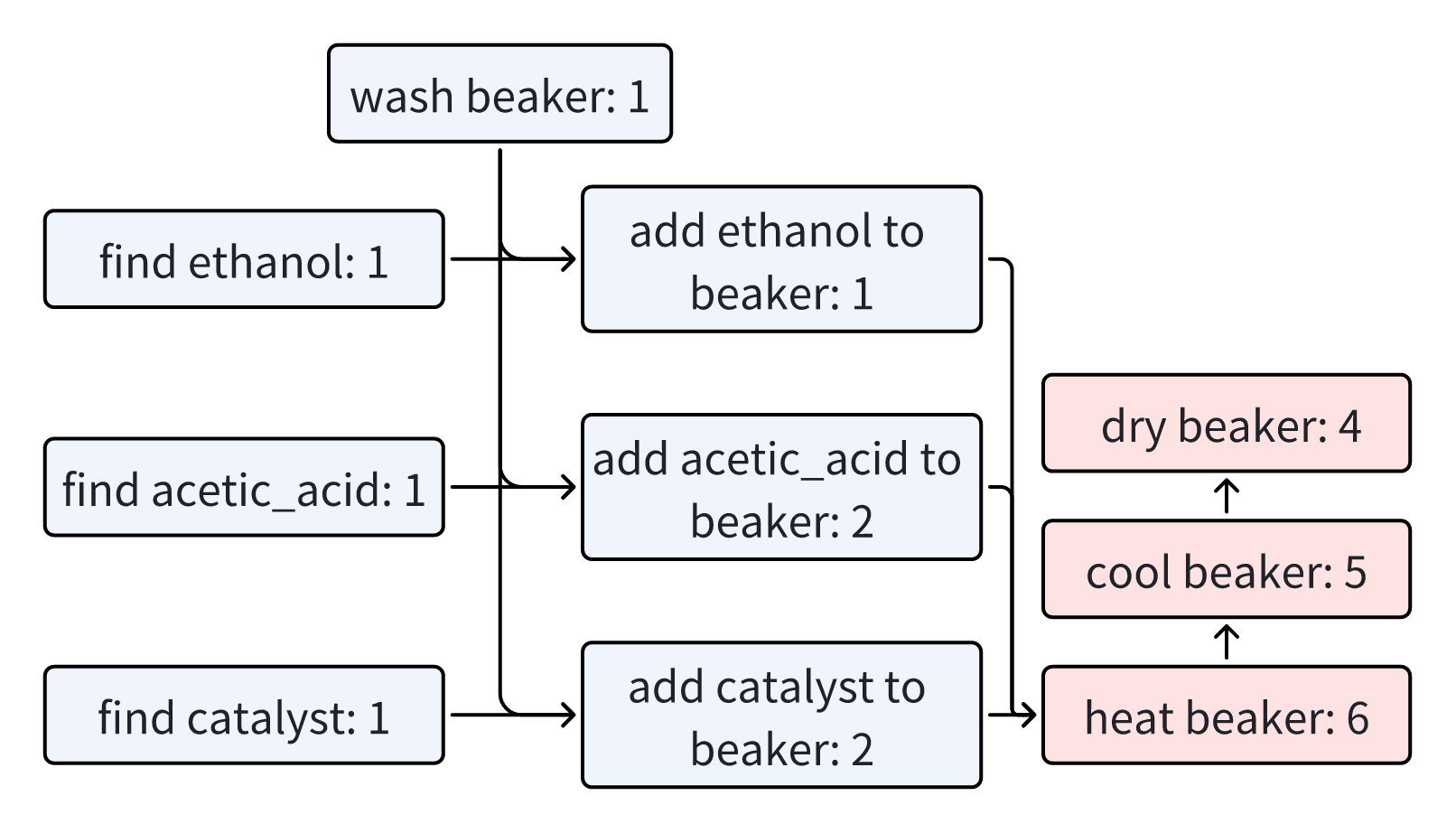}}
        \subfigure[The eighth task in the laboratory work scenario.] { \label{fig:l8}
		\includegraphics[width=0.44\linewidth]{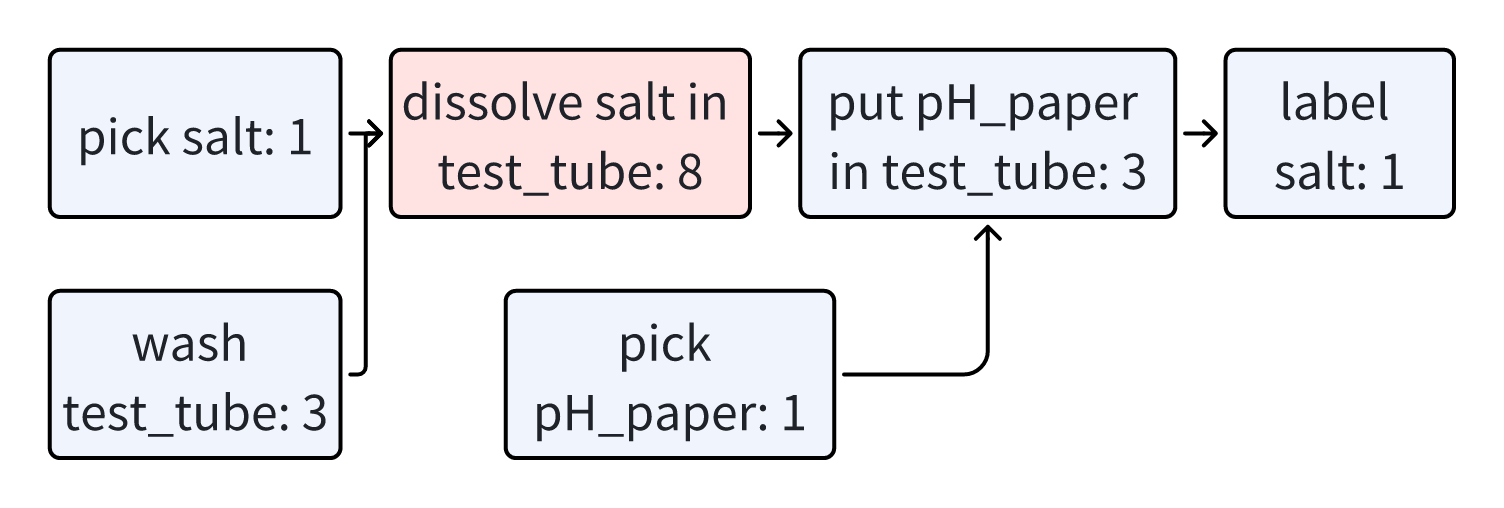}}
        \subfigure[The ninth task in the laboratory work scenario.] { \label{fig:l9}
		\includegraphics[width=0.44\linewidth]{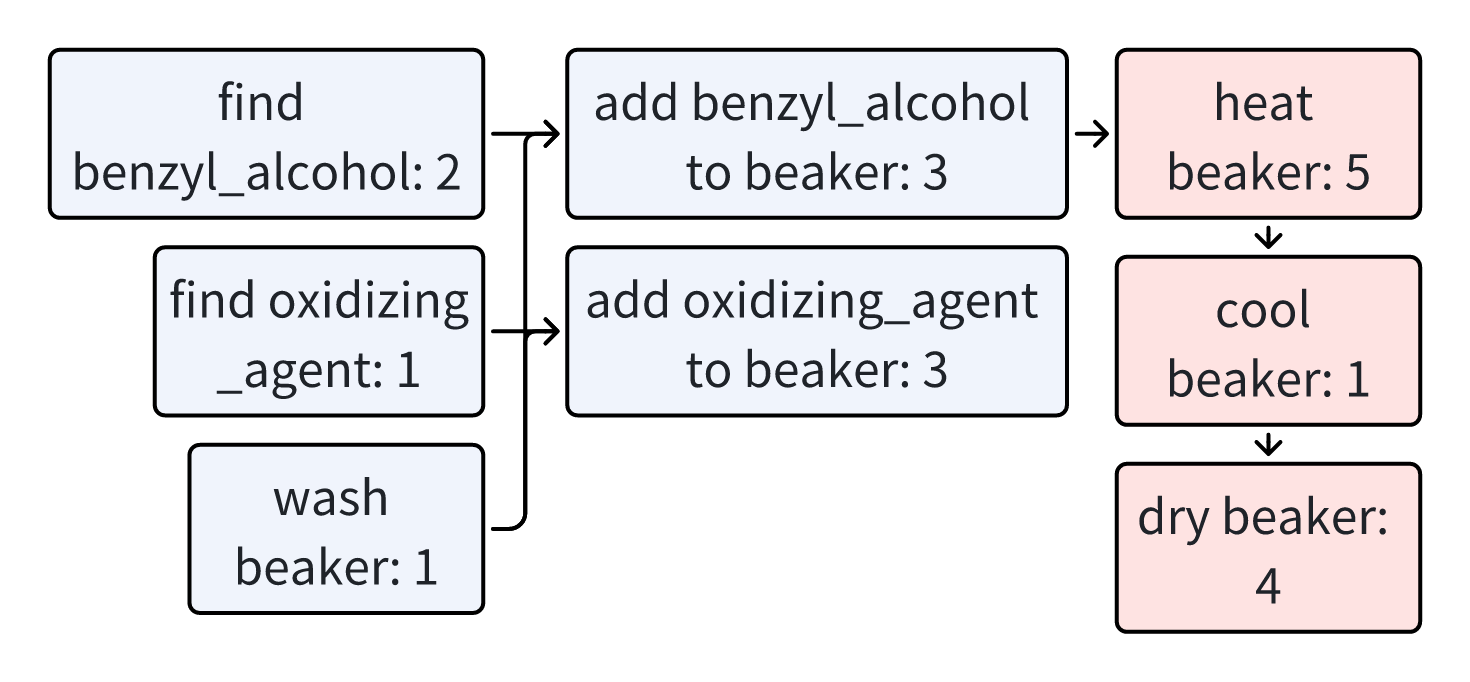}}
        \subfigure[The tenth task in the laboratory work scenario.] { \label{fig:l10}
		\includegraphics[width=0.44\linewidth]{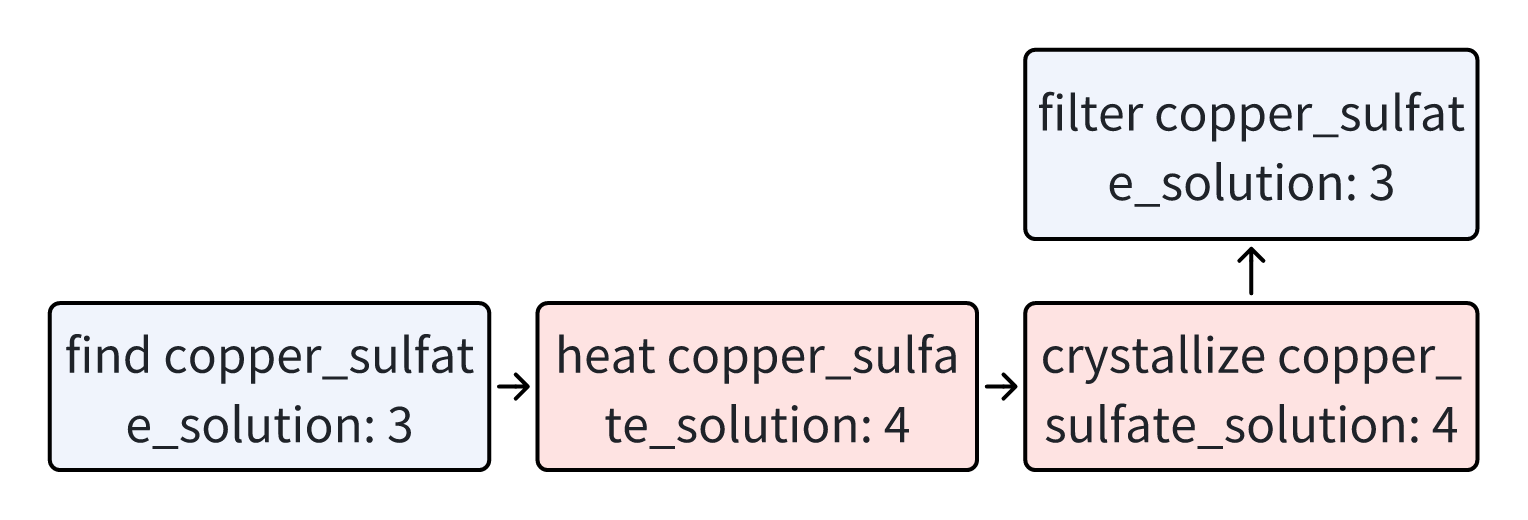}}
    \caption{The action dependencies and durations for the ten tasks in the laboratory work scenario. Actions that occupy the agent, preventing them from doing anything else, are indicated with a blue background. In contrast, actions that do not occupy the agent, allowing for parallel tasks, are marked with a red background.}
    \label{fig:laboratory_tasks}
\end{figure*}

\method contains 30 tasks in cooking, household activity, and laboratory work scenarios. 
To illustrate how to complete a task, we show the flow chart for each task in Figure~\ref{fig:cooking_tasks}, Figure~\ref{fig:householding_tasks} and Figure~\ref{fig:laboratory_tasks}.

\subsection{Actions}
\label{app:action}
The environment implements 45 actions, and each action has a description. 
We show the details of these actions in Table~\ref{tab:actions}. 
\begin{table*}[!ht]
\small
\begin{tabular}{ll}
\toprule
\textbf{Action} & \textbf{Description} \\
\midrule
\coloredtexttt{pick} \coloredtextttobj{OBJ} & Pick the unpicked item \\
\coloredtexttt{cook} \coloredtextttobj{OBJ1} \coloredtexttt{in} \coloredtextttobj{OBJ2} & Cook the raw item until it's cooked through \\
\coloredtexttt{chop} \coloredtextttobj{OBJ} & Chop the whole item into sliced pieces \\
\coloredtexttt{fry} \coloredtextttobj{OBJ1} \coloredtexttt{in} \coloredtextttobj{OBJ2} & Fry the raw item until it is fried to perfection \\
\coloredtexttt{wash} \coloredtextttobj{OBJ} & Wash the dirty item to make clean \\
\coloredtexttt{bake} \coloredtextttobj{OBJ1} \coloredtexttt{in} \coloredtextttobj{OBJ2} & Bake the raw item in the oven until it's roasted \\
\coloredtexttt{activate} \coloredtextttobj{OBJ} & Activate the inactive device to turn it active \\
\coloredtexttt{pour} \coloredtextttobj{OBJ1} \coloredtexttt{into} \coloredtextttobj{OBJ2} & Pour the liquid in item into the empty container until it is full \\
\coloredtexttt{brew} \coloredtextttobj{OBJ1} \coloredtexttt{with} \coloredtextttobj{OBJ2} & Brew the dry item leaves with the container until they're steeped \\
\coloredtexttt{gather} \coloredtextttobj{OBJ} & Gather the scattered items until it is collected \\
\coloredtexttt{scrape} \coloredtextttobj{OBJ1} \coloredtexttt{into} \coloredtextttobj{OBJ2} & Scrape the contents from the full item into th empty item \\
\coloredtexttt{place} \coloredtextttobj{OBJ1} \coloredtexttt{into} \coloredtextttobj{OBJ2} & Place the unplaced item into the right place \\
\coloredtexttt{fill} \coloredtextttobj{OBJ1} \coloredtexttt{with} \coloredtextttobj{OBJ2} & Fill the container with something \\
\coloredtexttt{hoe} \coloredtextttobj{OBJ} & Hoe the uncultivated item until it is cultivated and ready for planting \\
\coloredtexttt{weed\_with} \coloredtextttobj{OBJ} & Weed with the item \\
\coloredtexttt{set\_up} \coloredtextttobj{OBJ} &  Set up the item that is not set yet until it is already set \\
\coloredtexttt{iron} \coloredtextttobj{OBJ} & Iron the wrinkled item until they are smooth \\
\coloredtexttt{put} \coloredtextttobj{OBJ1} \coloredtexttt{on} \coloredtextttobj{OBJ2} & Put the item on the right place \\
\coloredtexttt{add} \coloredtextttobj{OBJ1} \coloredtexttt{to} \coloredtextttobj{OBJ2} & Add one item to the container \\
\coloredtexttt{rinse} \coloredtextttobj{OBJ} & Rinse the dry item \\
\coloredtexttt{find} \coloredtextttobj{OBJ} & Find the missed item so that it is found and can be used \\
\coloredtexttt{heat} \coloredtextttobj{OBJ} &  Heat the cool item until it is hot \\
\coloredtexttt{dilute} \coloredtextttobj{OBJ} & Dilute the concentrated item until it is diluted \\
\coloredtexttt{cut} \coloredtextttobj{OBJ} & Cut the whole item into divided pieces \\
\coloredtexttt{dissolve} \coloredtextttobj{OBJ1} \coloredtexttt{in} \coloredtextttobj{OBJ2} & Dissolve the solid item in the liquid until it is dissolved \\
\coloredtexttt{polish} \coloredtextttobj{OBJ} & Polish the rusty item until it is polished \\
\coloredtexttt{empty} \coloredtextttobj{OBJ} & Empty the full item until it is empty \\
\coloredtexttt{hanging} \coloredtextttobj{OBJ} & Hang the item \\
\coloredtexttt{water} \coloredtextttobj{OBJ1} \coloredtexttt{by} \coloredtextttobj{OBJ2} & Water the item by something \\
\coloredtexttt{trim} \coloredtextttobj{OBJ} & Trim the overgrown item \\
\coloredtexttt{plant} \coloredtextttobj{OBJ} & Plant the uncultivated item until it is planted \\
\coloredtexttt{store} \coloredtextttobj{OBJ} & Store the unstored item \\
\coloredtexttt{stir} \coloredtextttobj{OBJ1} \coloredtexttt{with} \coloredtextttobj{OBJ2} & Stir the separate liquid in item with something until it is homogeneous \\
\coloredtexttt{soak} \coloredtextttobj{OBJ1} \coloredtexttt{in} \coloredtextttobj{OBJ2} & Soak the dry item in something until it is wet \\
\coloredtexttt{mop} \coloredtextttobj{OBJ} & Mop the dirty item until it is clean \\
\coloredtexttt{read} \coloredtextttobj{OBJ} & Read the unknown item \\
\coloredtexttt{fold} \coloredtextttobj{OBJ} &  Fold the spread item until it is tidy \\
\coloredtexttt{crush} \coloredtextttobj{OBJ} & Crush the intact item until it is crushed \\
\coloredtexttt{cool} \coloredtextttobj{OBJ} & Cool the hot item until it is cool \\
\coloredtexttt{dry} \coloredtextttobj{OBJ} &  Dry the item until it is dry \\
\coloredtexttt{wipe} \coloredtextttobj{OBJ} & Wipe the dirty item until it is clean \\
\coloredtexttt{put} \coloredtextttobj{OBJ1} \coloredtexttt{in} \coloredtextttobj{OBJ2} &  Put the item in something \\
\coloredtexttt{label} \coloredtextttobj{OBJ} &  Give the ambiguous item a label \\
\coloredtexttt{crystallize} \coloredtextttobj{OBJ} &  Crystallize the fluid item until it is crystallized \\
\coloredtexttt{filter} \coloredtextttobj{OBJ} & Filter the mixed item until it is refined \\
\bottomrule
\end{tabular}
\caption{Details of actions with descriptions.}
\label{tab:actions}
\end{table*}

\subsection{Action Types}
\label{app:actiontype}
\definecolor{mygray}{RGB}{240, 240, 240}

\begin{table*}[t]
\centering
\small
\begin{tabular}{ccll}
\toprule
\textbf{Type} & \textbf{Subtype} & \textbf{Explanation} &  \textbf{Example: Make tea}\\
\midrule

\multicolumn{1}{c}{\multirow{10}[6]{*}{\makecell{Incorrect\\Actions}}} & {\color{red}Invalid Action/Object} & \makecell[l]{An action does not in the \\action space or non-existent \\objects are visited.} & \multirow{8}[10]{*}{\makecell[l]{<Valid Actions>\\activate; wash; brew with; pour into\\<Objects>\\tea(dry); kettle(inactive); \\teapot(dirty); cup(dirty) \\ <Trajectory> \\
T=1: {\color{red}clean teapot} \\ T=2: {\color{violet}brew tea with teapot} \\ T=3: {\color{teal}wash teapot} \\ T=4: {\color{blue}wash kettle} \\ T=5: {\color{brown}{wash teapot}} \\T=6: {\color{teal}activate kettle} \\ T=7: {\color{gray}wait} \\ ... }} \\
\cmidrule{2-3}   & {\color{brown}Repeating Completed Action} & \makecell[l]{An action is in the action space and \\matches the objects, but it \\has already been completed.} &  \\ 
\cmidrule{2-3}  & {\color{violet}Dependency Violation} & \makecell[l]{An action is in the action space \\and matches the objects, but the \\necessary prerequisite actions \\have not been completed.} &  \\
\cmidrule{2-3}  & {\color{blue}Object-Mismatched Action} & \makecell[l]{An action is in the action space \\and the object is available, \\but they do not match.} &   \\
\cmidrule{1-3}
\multicolumn{1}{c}{\multirow{3}[6]{*}{\makecell{Correct\\Actions}}} & {\color{teal}Valid Action} & \makecell[l]{An action is in the action space \\and matches the objects.} &  \\
\cmidrule{2-3} & {\color{gray}Wait} & \makecell[l]{An action is used to pass the \\current time.} &   \\
\bottomrule
\end{tabular}
\caption{Action types and their explanations with an example.}
\label{tab:action_type} 
\end{table*}

As shown in Table~\ref{tab:action_type}, we define 4 incorrect action types and 2 correct action types for analyzing why agents fail in multitasking.

\subsection{Greedy Algorithm}
\label{app:algo}
\begin{algorithm}[t]
    \caption{Greedy Algorithm for Minimal Time Calculation}
    \label{alg:minimal_time}
    \KwIn{Set of actions $\mathcal{A}$, Durations $\mathcal{T}$, Dependencies $p(\mathcal{A})$.}
    \KwOut{Minimal time $\mathcal{T}_{\text{min}}$.}
    \BlankLine
    Define non-occupied actions $\mathcal{A*}$ and occupied actions $\mathcal{A'}$ from $\mathcal{A}$.
    
    Sort $\mathcal{A*}$ by $\mathcal{T}$ in descending order.
    
    $\mathcal{A} \leftarrow \text{concatenate}(\mathcal{A*}, \mathcal{A'})$.
    
    Initialize Action\_list as an empty list.
    
    \ForEach{$a_i \in \mathcal{A}$}{
        $P \leftarrow \text{BFS}(a_i, p(a_i))$ to collect prerequisites.

        \ForEach{$p_i \in P$}{
            \If{$p_i \in \mathcal{A}$}{
                Action\_list.append($p_i$).
                
                Remove $p_i$ from $\mathcal{A}$.
            }
        }
        Action\_list.append($a_i$).
    }
    
    $\mathcal{T}_{\text{min}} \leftarrow 0$.
    
    \While{not empty $\mathcal{A*}$ or $\mathcal{A'}$}{
            \ForEach{$a_i \in$ Action\_list}{
                \If{check\_dependency($a_i$)}{
                    \eIf{$a_i \in \mathcal{A*}$}{
                        $\mathcal{T}_{\text{min}} \leftarrow \mathcal{T}_{\text{min}} + 1$.
                        
                        Remove $a_i$ from $\mathcal{A*}$.
                    }{
                        $\mathcal{T}_{\text{min}} \leftarrow \mathcal{T}_{\text{min}} + \mathcal{T}(a_i)$.
                        
                        Remove $a_i$ from $\mathcal{A'}$.
                    }
                    break.
                }
            }
        Increment $\mathcal{T}_{\text{min}}$ by 1 if no action is performed.
    }
\end{algorithm}

We show the greedy algorithm in Algorithm~\ref{alg:minimal_time}.

\section{Examples of \method}

\subsection{Tasks}
\label{app:example_task}
Table~\ref{list:task1}, \ref{list:task2} and \ref{list:task3} present some examples of task combinations in \method for a better understanding.
\subsection{Interaction}
\label{app:example_inter}
Table~\ref{list:task1} shows an example of interaction between an agent and environment in the cooking scenario.
\subsection{Self-plan}
\label{app:example_sp}
Table~\ref{list:selfplan} shows the prompt of the self-plan method.

\lstset{
    backgroundcolor=\color[RGB]{250,250,250},
    breaklines=true,
    breakindent=0pt,
    basicstyle=\ttfamily\small,
}
\begin{table*}[!ht]
\begin{lstlisting}
As an AI agent, your objective is to efficiently complete a series of tasks as described. You must adhere to the specific requirements and constraints of each task, including dependencies and timing. Efficiency is key; complete all tasks in the shortest possible time. I will provide instructions regarding actions and objects.

**Action Protocol**:
- You can perform only one action at a time.
- After each observation from the environment, output an action based on that observation and the instructions.
- Actions fall into two categories:
- Continuous Actions: Perform these actions until completion (e.g., "wash OBJ").
- Autonomous Actions: These progress over time, allowing simultaneous tasks (e.g., "heat OBJ").
- Follow the "Valid Actions" format for your output (e.g., "wash cup").
- If no action is required, use "wait" to skip the current time.
- Output the action explicitly (e.g., "wash cup").
- Select object names (OBJ) from the list of Available Objects (e.g., use "rice" instead of "cooked rice").

**Task 1**
<Description>
- Prepare a noodle dish, which consists of cooked noodle, fried mushrooms and shrimp.
<Valid Actions and Usages>
- pick OBJ: Pick the unpicked item.
- cook OBJ1 in OBJ2: Cook the raw item until it's cooked through.
- chop OBJ: Chop the whole item into sliced pieces.
- fry OBJ1 in OBJ2: Fry the raw item until it is fried to perfection.
- add OBJ1 to OBJ2: Add one item to the container.
- wash OBJ: Wash the dirty item to make clean.
- wait: pass the current time without doing anything.

**All Available Objects (OBJ)**
noodle; mushroom; shrimp; fryer; pot; dish

**The Initial States of Objects**
noodle: unpicked; mushroom: unpicked; shrimp: unpicked; fryer: empty; pot: empty; dish: dirty
\end{lstlisting}
\caption{An example of \# Task=1 scenario.}
\label{list:task1}
\end{table*}
\lstset{
    backgroundcolor=\color[RGB]{250,250,250},
    breaklines=true,
    breakindent=0pt,
    basicstyle=\ttfamily\small,
}
\begin{table*}[!ht]
\begin{lstlisting}
As an AI agent, your objective is to efficiently complete a series of tasks as described. You must adhere to the specific requirements and constraints of each task, including dependencies and timing. Efficiency is key; complete all tasks in the shortest possible time. I will provide instructions regarding actions and objects.

**Action Protocol**:
- You can perform only one action at a time.
- After each observation from the environment, output an action based on that observation and the instructions.
- Actions fall into two categories:
- Continuous Actions: Perform these actions until completion (e.g., "wash OBJ").
- Autonomous Actions: These progress over time, allowing simultaneous tasks (e.g., "heat OBJ").
- Follow the "Valid Actions" format for your output (e.g., "wash cup").
- If no action is required, use "wait" to skip the current time.
- Output the action explicitly (e.g., "wash cup").
- Select object names (OBJ) from the list of Available Objects (e.g., use "rice" instead of "cooked rice").

**Task 1**
<Description>
- Prepare and bake a cheese and tomato pizza
<Valid Actions and Usages>
- pick OBJ: Pick the unpicked item.
- chop OBJ: Chop the whole item into sliced pieces.
- wash OBJ: Wash the dirty item to make clean.
- add OBJ1 to OBJ2: Add one item to the container.
- bake OBJ1 in OBJ2: Bake the raw item in the oven until it's roasted.
- wait: pass the current time without doing anything.

**Task 2**
<Description>
- Prepare chicken and potato stir-fry, which consists of fried chicken and fried potato.
<Valid Actions and Usages>
- pick OBJ: Pick the unpicked item.
- chop OBJ: Chop the whole item into sliced pieces.
- fry OBJ1 in OBJ2: Fry the raw item until it is fried to perfection.
- add OBJ1 to OBJ2: Add one item to the container.
- wash OBJ: Wash the dirty item to make clean.
- wait: pass the current time without doing anything.

**All Available Objects(OBJ)**
dish_1; dish_2; dough; cheese; tomato; oven; chicken; potato; fryer

**The Initial States of Objects**
dish_1: dirty; dish_2: dirty; dough: unpicked; cheese: unpicked; tomato: unpicked; oven: empty; chicken: unpicked; potato: unpicked; fryer: empty
\end{lstlisting}
\caption{An example of \# Task=2 scenario.}
\label{list:task2}
\end{table*}
\lstset{
    backgroundcolor=\color[RGB]{250,250,250},
    breaklines=true,
    breakindent=0pt,
    basicstyle=\ttfamily\small,
}
\begin{table*}[!ht]
\begin{lstlisting}
As an AI agent, your objective is to efficiently complete a series of tasks as described. You must adhere to the specific requirements and constraints of each task, including dependencies and timing. Efficiency is key; complete all tasks in the shortest possible time. I will provide instructions regarding actions and objects.

**Action Protocol**:
- You can perform only one action at a time.
- After each observation from the environment, output an action based on that observation and the instructions.
- Actions fall into two categories:
- Continuous Actions: Perform these actions until completion (e.g., "wash OBJ").
- Autonomous Actions: These progress over time, allowing simultaneous tasks (e.g., "heat OBJ").
- Follow the "Valid Actions" format for your output (e.g., "wash cup").
- If no action is required, use "wait" to skip the current time.
- Output the action explicitly (e.g., "wash cup").
- Select object names (OBJ) from the list of Available Objects (e.g., use "rice" instead of "cooked rice").

**Task 1**
<Description>
- Prepare a garden bed for planting flowers by using sprinkling can filled with herbicide, hoeing, and weeding
<Valid Actions and Usages>
- add OBJ1 to OBJ2: Add one item to the container.
- weed_with OBJ: Weed with the item.
- hoe OBJ: Hoe the uncultivated item until it is cultivated and ready for planting.
- plant OBJ: Plant the uncultivated item until it is planted
- wait: pass the current time without doing anything.

**Task 2**
<Description>
- Iron a suit and store it properly
<Valid Actions and Usages>
- set_up OBJ: Set up the item that is not set yet until it is already set.
- put OBJ1 on OBJ2: Put the item on the right place.
- heat OBJ: Heat the cool item until it is hot.
- iron OBJ: Iron the wrinkled item until they are smooth.
- store OBJ: Store the unstored item\nwait: pass the current time without doing anything.

**Task 3**
<Description>
- Make a cup of coffee
<Valid Actions and Usages>
- add OBJ1 to OBJ2: Add one item to the container.
- activate OBJ: Activate the inactive device to turn it active.
- wash OBJ: Wash the dirty item to make clean.
- pour OBJ1 into OBJ2: Pour the liquid in item into the empty container until it is full.
- wait: pass the current time without doing anything.

**All Available Objects(OBJ)**
sprinkling_can; herbicide; land; flower; ironing_board; suit; iron; coffee_beans; coffee_machine; water; cup

**The Initial States of Objects**
sprinkling_can: empty; herbicide: not added; land: uncultivated; flower: uncultivated; ironing_board: not set yet; suit: not put on right place; iron: cool; coffee_beans: not added; coffee_machine: empty; water: not added; cup: dirty
\end{lstlisting}
\caption{An example of \# Task=3 scenario.}
\label{list:task3}
\end{table*}

\lstset{
    backgroundcolor=\color[RGB]{250,250,250},
    breaklines=true,
    breakindent=0pt,
    basicstyle=\ttfamily\small,
}
\begin{table*}[!ht]
\begin{lstlisting}
<|Environment|>:
As an AI agent, your objective is to efficiently complete a series of tasks as described. You must adhere to the specific requirements and constraints of each task, including dependencies and timing. Efficiency is key; complete all tasks in the shortest possible time. I will provide instructions regarding actions and objects.

**Action Protocol**:
- You can perform only one action at a time.
- After each observation from the environment, output an action based on that observation and the instructions.
- Actions fall into two categories:
- Continuous Actions: Perform these actions until completion (e.g., "wash OBJ").
- Autonomous Actions: These progress over time, allowing simultaneous tasks (e.g., "heat OBJ").
- Follow the "Valid Actions" format for your output (e.g., "wash cup").
- If no action is required, use "wait" to skip the current time.
- Output the action explicitly (e.g., "wash cup").
- Select object names (OBJ) from the list of Available Objects (e.g., use "rice" instead of "cooked rice").

**Task 1**
<Description>
- Make a dish of beef fried rice, which consists of cooked rice and fried beef.
<Valid Actions and Usages>
- wash OBJ: Wash the dirty item to make clean.
- pick OBJ: Pick the unpicked item.
- cook OBJ1 in OBJ2: Cook the raw item until it's cooked through.
- chop OBJ: Chop the whole item into sliced pieces.
- fry OBJ1 in OBJ2: Fry the raw item until it is fried to perfection.
- add OBJ1 to OBJ2: Add one item to the container.
- wait: pass the current time without doing anything.

**All Available Objects(OBJ)**
rice; beef; pot; fryer; dish

**The Initial States of Objects**
rice: unpicked; beef: unpicked; pot: empty; fryer: empty; dish: dirty

<|Environment|>:
In t=1, your action is: 

<|Agent|>:
pick dish

<|Environment|>:
You are doing ``wash dish``, it will take 2 minutes.
rice is picked.

In t=3, your action is: 

<|Agent|>:
cook rice in pot

<|Environment|>:
You are doing ``cook rice in pot``, it will take 4 minutes.

In t=4, your action is: 

<|Agent|>:
wait

<|Environment|>:
You wait for one minute.

...
\end{lstlisting}
\caption{An example of interaction in cooking scenario.}
\label{list:task4}
\end{table*}

\lstset{
    backgroundcolor=\color[RGB]{250,250,250},
    breaklines=true,
    breakindent=0pt,
    basicstyle=\ttfamily\small,
}
\begin{table*}[!ht]
\begin{lstlisting}
<|Environment|>:
As an AI agent, your objective is to efficiently complete a series of tasks as described. You must adhere to the specific requirements and constraints of each task, including dependencies and timing. Efficiency is key; complete all tasks in the shortest possible time. I will provide instructions regarding actions and objects.

**Action Protocol**:
- You can perform only one action at a time.
- After each observation from the environment, output an action based on that observation and the instructions.
- Actions fall into two categories:
- Continuous Actions: Perform these actions until completion (e.g., "wash OBJ").
- Autonomous Actions: These progress over time, allowing simultaneous tasks (e.g., "heat OBJ").
- Follow the "Valid Actions" format for your output (e.g., "wash cup").
- If no action is required, use "wait" to skip the current time.
- Output the action explicitly (e.g., "wash cup").
- Select object names (OBJ) from the list of Available Objects (e.g., use "rice" instead of "cooked rice").

**Task 1**
<Description>
- Make a dish of beef fried rice, which consists of cooked rice and fried beef.
<Valid Actions and Usages>
- wash OBJ: Wash the dirty item to make clean.
- pick OBJ: Pick the unpicked item.
- cook OBJ1 in OBJ2: Cook the raw item until it's cooked through.
- chop OBJ: Chop the whole item into sliced pieces.
- fry OBJ1 in OBJ2: Fry the raw item until it is fried to perfection.
- add OBJ1 to OBJ2: Add one item to the container.
- wait: pass the current time without doing anything.

**All Available Objects(OBJ)**
rice; beef; pot; fryer; dish

**The Initial States of Objects**
rice: unpicked; beef: unpicked; pot: empty; fryer: empty; dish: dirty

Given  the list of valid actions, available objects, and the task descriptions (goal), please perform the following steps:
- Identify and list all of the necessary actions required to accomplish the task's goal.
- For each action, determine and note the specific objects that are required.
- Assess and map out any dependencies between actions, indicating which actions must precede others.
- Arrange the actions in a logical sequence that respects the dependencies and leads efficiently towards completing the task.
- If any action has multiple dependencies, list them in order of priority based on the task's constraints and goal.
- Present the final action sequence in a clear and ordered list, ensuring that the progression of steps will achieve the task's objective.

The key to efficiency:
- When completing tasks, some actions are non-occupied actions(Type 2), meaning you can perform other actions simultaneously.
- To maximize efficiency, adhere to the following principle: always start the non-occupied action you anticipate will be the most time-consuming as early as possible. 
- You should perform actions during idle times as much as possible to minimize the time spent doing nothing.

...
\end{lstlisting}
\caption{Prompt of self-plan method in cooking scenario.}
\label{list:selfplan}
\end{table*}

\end{document}